\documentclass{article}

\usepackage[preprint]{neurips_2025}

\usepackage[utf8]{inputenc} 
\usepackage[T1]{fontenc}    
\usepackage{hyperref}       
\usepackage{url}            
\usepackage{booktabs}       
\usepackage{amsfonts}       
\usepackage{nicefrac}       
\usepackage{microtype}      
\usepackage{xcolor}         
\usepackage{microtype}
\usepackage{graphicx}
\usepackage{subfigure}
\usepackage{booktabs}
\usepackage{amsmath}
\usepackage{amssymb}
\usepackage{mathtools}
\usepackage{amsthm}
\usepackage{wrapfig}
\usepackage{listings}
\usepackage[capitalize,noabbrev]{cleveref}

\theoremstyle{plain}
\newtheorem{theorem}{Theorem}[section]

\theoremstyle{definition}

\newtheorem{assumption}[theorem]{Assumption}
\theoremstyle{remark}

\usepackage{wrapfig}
\usepackage{algorithm}
\usepackage{algorithmic}
\usepackage{enumitem}
\usepackage{graphicx}
\usepackage{amssymb}
\usepackage{mathrsfs}
\usepackage{amsmath}
\usepackage{mathtools}
\usepackage{subfigure}
\usepackage{graphicx}
\usepackage{wrapfig}
\usepackage{multirow}
\usepackage{natbib}
\usepackage{tabularx}

\usepackage{enumitem}
\usepackage{colortbl}
\usepackage{bbm}
\usepackage{xspace}
\usepackage[normalem]{ulem}
\useunder{\uline}{\ul}{}
\definecolor{codegray}{rgb}{0.5,0.5,0.5}
\definecolor{codepurple}{rgb}{0.58,0,0.82}
\definecolor{backcolour}{rgb}{0.95,0.95,0.92}
\usepackage[textsize=tiny]{todonotes}

\title{Efficient Unsupervised Environment Design through Hierarchical Policy Representation Learning}

\author{%
  Dexun Li\thanks{Equal contribution.}  \qquad \qquad 
  Sidney Tio\footnotemark[1] \qquad \qquad 
  Pradeep Varakantham \\
  School of Computing and Information Systems \\
  Singapore Management University \\
  \texttt{dexunli.2019@phdcs.smu.edu.sg}
}

\begin{document}

\maketitle

\begin{abstract}
Unsupervised Environment Design (UED) has emerged as a promising approach to developing general-purpose agents through automated curriculum generation. Popular UED methods focus on Open-Endedness, where teacher algorithms rely on stochastic processes for infinite generation of useful environments. This assumption becomes impractical in resource-constrained scenarios where teacher-student interaction opportunities are limited. To address this challenge, we introduce a hierarchical Markov Decision Process (MDP) framework for environment design. Our framework features a teacher agent that leverages student policy representations derived from discovered evaluation environments, enabling it to generate training environments based on the student's capabilities. To improve efficiency, we incorporate a generative model that augments the teacher's training dataset with synthetic data, reducing the need for teacher-student interactions. In experiments across several domains, we show that our method outperforms baseline approaches while requiring fewer teacher-student interactions in a single episode. The results suggest the applicability of our approach in settings where training opportunities are limited. The code is available at https://github.com/HughLee1994/Efficient-UED-through-Hierarchical-Policy-Representation-Learning.
\end{abstract}

\section{Introduction}
Reinforcement Learning (RL, \citet{sutton1998introduction}) has advanced the development of autonomous agents capable of accomplishing complex tasks. A key challenge in this field is to train agents that perform well in previously unseen environments, known as zero-shot transfer performance. Unsupervised Environment Design (UED, \citet{dennis2020emergent}) has emerged as a promising approach to address this challenge. UED is a RL training framework that automatically generates and evolves training environments to challenge learning agents without human supervision. Instead of relying on manually crafted scenarios, UED creates a co-evolutionary process where an environment generator and a student agent develop in parallel – as the agent masters certain challenges, the environment generator responds by creating increasingly complex scenarios. This automated approach helps develop a more robust student agent by exposing them to a diverse range of training experiences while eliminating the need for human designers to manually create training curricula.

\begin{figure}[t]
\vskip 0.2in
\begin{center}
\centerline{\includegraphics[width=0.8\linewidth]{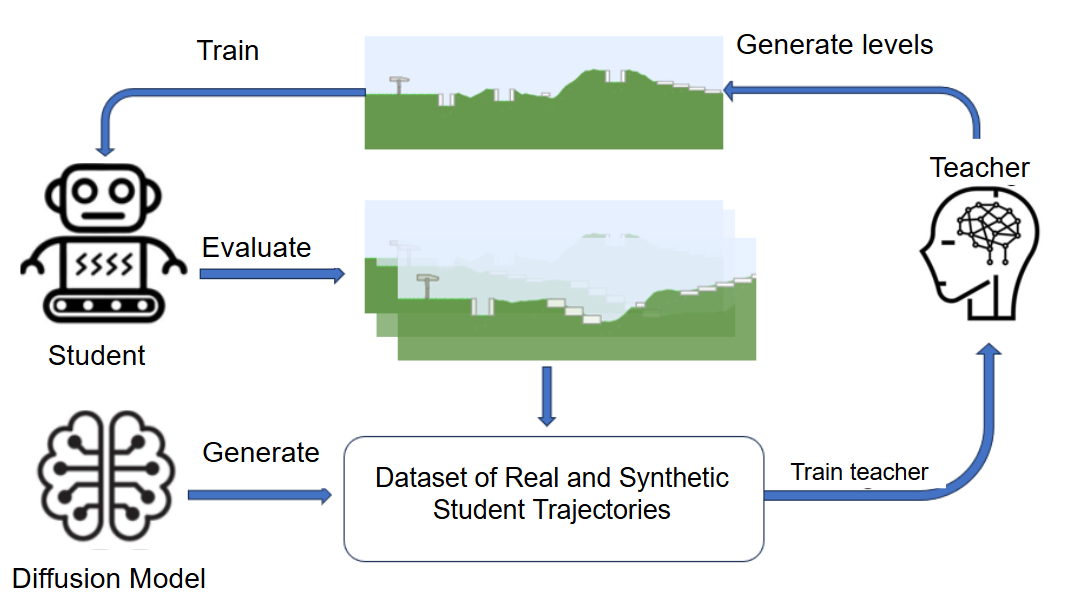}}
\caption{The overall framework of SHED. SHED uses student's performance on select evaluation environments as its state to suggest the next appropriate challenge for students to train in.}
\label{fig:framework}
\end{center}
\vskip -0.2in
\end{figure}

While UED methods have shown promising zero-shot transfer performance, they face critical limitations: inefficient random exploration strategies in large design spaces (e.g. \citet{costales2024enabling,parker2022evolving,Li_Li_Varakantham_2025}. In general, these methods do not discuss training horizons as a constraint. Our work addresses the fundamental challenge of developing student agents with generalized capabilities under strict environmental budgets and practical training constraints, requiring a sophisticated teacher policy calibrated to students' skill levels.
Unlike previous UED approaches focused on open-ended learning, we target efficient student development within limited environmental budgets—particularly valuable when training multiple student agents or when teacher-student interactions are costly. In our work, the initial investment in teacher training is amortized across multiple student sessions, increasing efficiency with each additional student trained.

We present a novel adaptive environment design framework using a hierarchical Markov Decision Process (MDP), where a teacher agent generates training environments for a student agent. The teacher approximates the student's capabilities through performance evaluations across diverse environments, using this as state information for deciding next training environments. To overcome the computational burden of collecting teacher experience, where each transition requires complete student training, we employ diffusion models ~\citep{saharia2022photorealistic, sohl2015deep,ho2020denoising} to generate synthetic student policy evolution trajectories set. Our method, called \emph{S}ynthetically-enhanced \emph{H}ierarchical \emph{E}nvironment \emph{D}esign (SHED), efficiently creates environments of progressive complexity matched to student capabilities.


Our key contributions are:
\begin{itemize}
    \item A novel hierarchical MDP framework for UED that introduces a straightforward, yet effective, method to represent the student agent's current capability level
    \item SHED, which employs diffusion-based techniques to generate synthetic experiences, accelerating off-policy teacher agent training
    \item Empirical evidence demonstrating our method's superior performance over existing UED approaches across different task domains, achieving better general capability under resource constraints
\end{itemize}

\section{Related Work}
\paragraph{Unsupervised Environment Design}

The goal of UED is to generate environmental sequences that develop generally capable student agents. UED approaches fall into two main categories: RL-based and domain randomization-based environment generation.

The RL-based approach \citep{dennis2020emergent,azad2023clutr,li2023diversity} was first formalized as a self-supervised RL paradigm for environment generation. In general, these methods feature an RL teacher that receives feedback on student performance, with the goal of developing a successful teacher that adapts environment challenges as the student policy improves over training.

In contrast, Domain Randomization (DR; \cite{tobin2017domain}) achieves strong results by training agents in randomly generated environments without complex RL teachers. Two notable advances include \citet{jiang2021prioritized}, which introduces an emergent curriculum by prioritizing high-regret environments, and ACCEL \citep{parker2022evolving}, which employs a principled curriculum of increasingly complex environments through random modifications. Given ACCEL's leading performance, we use it and a variant as our primary baselines.

DR remains foundational for sim-to-real transfer \citep{yuan2024learningmanipulateanywherevisual,duan2024learningvisionbasedbipedallocomotion,da2025survey}, training policies in simulated environments that ideally performs as well in real-world scenarios. Our work bridges UED and sim-to-real approaches by focusing on methods applicable in both domains, particularly addressing realistic time and resource constraints.

Current UED methods assume unrestricted student-teacher interactions without considering horizon constraints, creating an idealized scenario that limits practical applications. Our work directly addresses these limitations to enhance UED's applicability in resource-constrained settings.

\paragraph{Generative Models in UED}
Recent UED advances leverage generative models, with \citet{tio2023transferable} using an Item Response Theory-based VAE \citep{kingma2013auto} to match student abilities with appropriate environments, and \citet{azad2023clutr} employing VAEs to compress complex environments into manageable representations for RL teachers. \citet{garcin2024dred} learns a VAE-based model that interpolates at the environment embedding level to discover new environments.

While these approaches focus on modeling environments, our work takes a different direction by applying generative models to represent student policies themselves. This captures the evolving capabilities of student agents comprehensively, enabling RL teachers to better understand and guide student development throughout the training process.

\section{Preliminaries}
\subsection{Underspecified Partially Observable MDP} \label{subsec:ued}
\citet{dennis2020emergent} first modelled UED with an Underspecified Partially Observable Markov Decision Process (UPOMDP), which is a tuple
$$\mathcal{M}=<A, O,\Theta, S^{\mathcal{M}},\mathcal{P}^{\mathcal{M}},\mathcal{I}^{\mathcal{M}},\mathcal{R}^{\mathcal{M}},\gamma>$$
\noindent The UPOMDP has a set $\Theta$ representing the free parameters of the environments, which are determined by the teacher agent and can be distinct to generate the next new environment. Further, these parameters are incorporated into the environment-dependent transition function $\mathcal{P}^{\mathcal{M}}: S \times A \times \Theta \rightarrow S$. Here $A$ represents the set of actions, $S$ is the set of states. Similarly, $\mathcal{I}^{\mathcal{M}}:S \rightarrow O$ is the environment-dependent observation function, $\mathcal{R}^{\mathcal{M}}$ is the reward function, and $\gamma$ is the discount factor. Specifically, given the environment parameters $\vec{\theta}\in\Theta$, we denote the corresponding environment instance as $\mathcal{M}_{\vec{\theta}}$. The student policy $\pi$ is trained to maximize the cumulative rewards $V^{\mathcal{M}_{\vec{\theta}}}(\pi) = \sum_{t=0}^T \gamma^t r_t$ in the given environment $\mathcal{M}_{\vec{\theta}}$ under a time horizon $T$, and $r_t$ are the collected rewards in $\mathcal{M}_{\vec{\theta}}$.

\subsection{Diffusion Probabilistic Models}\label{subsec:diff}
Diffusion models~\citep{sohl2015deep} are generative models that learn data distributions by defining two Markov chains: a forward process that gradually adds noise to data, and a learned reverse process that transforms noise back to data. Recent advances have demonstrated their effectiveness across various domains including time series forecasting~\citep{tashiro2021csdi}, robust learning~\citep{nie2022diffusion}, anomaly detection~\citep{wyatt2022anoddpm}, and text-to-image synthesis~\citep{nichol2021glide,saharia2022photorealistic}.

The forward process perturbs data $x_0$ with Gaussian noise over $K$ steps according to variance schedule $\{\beta_k\}_{k=1}^K$, ultimately converging to standard Gaussian noise. The reverse process uses neural networks to parameterize transition kernels $p_{\phi}(x_{k-1}|x_k)$ that progressively denoise samples. Training optimizes:

\begin{equation}
    \mathcal{L}(\phi) = \mathbb{E}_{x_0,k,{\epsilon}} \left[ \lVert {\epsilon}  - {\epsilon}_{\phi}(\sqrt{\bar{\alpha}_k}x_0 + \sqrt{1-\bar{\alpha}_k }{\epsilon}, k)  \rVert^2  \right]
\end{equation}

where ${\epsilon}_{\phi}$ predicts noise added to $x_0$. Upon training, new samples are generated by sampling Gaussian noise and iteratively applying the learned reverse process. See Appendix \ref{app:preliminary} for complete mathematical details of the forward and reverse chains.

\section{Approach}\label{sec:method}
\paragraph{Overview}
This section introduces \emph{S}ynthetically-enhanced \emph{H}ierarchical \emph{E}nvironment \emph{D}esign ({\em SHED}), our novel framework for Unsupervised Environment Design under resource constraints. SHED advances beyond existing approaches through two key innovations: (1) a hierarchical MDP formalism that enables the RL teacher to generate environments matched to the student's developmental stage using observed policy representations rather than random sampling, and (2) a generative model that creates synthetic training trajectories for off-policy teacher training, dramatically reducing direct student data collection requirements. Figure~\ref{fig:framework} illustrates our framework, with Algorithm~\ref{alg} providing the implementation details.

In practice, SHED operates in two distinct phases: an initial training phase that co-trains the teacher policy and generative model using our hierarchical MDP framework with limited student interactions; followed by a deployment phase where the trained teacher efficiently develops new student agents with minimal environmental exposure.



\begin{algorithm}[t]
\caption{\textit{SHED}}
\label{alg}
\begin{algorithmic}[1]
\REQUIRE 
   real data ratio $\psi\in [0,1]$, evaluate environment set $\bf{\theta}^{eval}$, reward function $R$; 
\STATE \textbf{Initialize:} diffusion model $D$, teacher policy $\Lambda$, real and synthetic replay buffer $\bf{\mathcal{B}}_{real},\bf{\mathcal{B}}_{syn}=\emptyset$;
\FOR{episode $ep=1,\dots,K$}
\STATE Initialize student policy $\pi$
\STATE Evaluate $\pi$ on $\bf{\theta}^{eval}$ and get state $s^u = p(\pi)$
\FOR{Budget $t=1,\dots,T$}
    \STATE generate $\vec{\theta} \sim \Lambda$, and create $\mathcal{M}_{\vec{\theta}}(\pi)$
    \STATE train $\pi$ on $\mathcal{M}_{\vec{\theta}}$ to maximize $V^{\vec{\theta}}(\pi)$
    \STATE evaluate $\pi$ on $\bf{\theta}^{eval}$ and get next state $s^\prime$
    \STATE compute teacher's reward $r_t$ according to $R$
    \STATE add experience $(s^u_t,\vec{\theta}, r_t^u,s^{u,\prime}_t)$ to ${\mathcal{B}}_{real}$
    \STATE train $D$ with samples from $\bf{\mathcal{B}}_{real}$ 
    \STATE generate synthetic experiences from $D$ and add them to $\bf{\mathcal{B}}_{syn}$
    \STATE train $\Lambda$ on samples from $\bf{\mathcal{B}}_{real} \bigcup {\mathcal{B}}_{syn}$ mixed with ratio $\psi$
    \STATE set $s= s^\prime$;
\ENDFOR
\ENDFOR
\ENSURE $\Lambda$,  $\pi$,  $D$
\end{algorithmic}
\end{algorithm}

\subsection{Hierarchical Environment Design}
The core challenge is designing a teacher algorithm that maximizes student agent performance under strict interaction budgets. SHED adopts an RL-based approach inspired by PAIRED~\citep{dennis2020emergent}, but introduces a hierarchical MDP framework with an upper-level teacher policy $\Lambda$ and lower-level student policy $\pi$.

Unlike existing methods that rely solely on regret feedback, SHED captures student policy nuances by mapping $\Lambda:{\Pi} \rightarrow \Theta$ from policy space to environment parameters. We represent the student policy $\pi$ as a performance vector $p(\pi)$ across diverse evaluation environments, providing a practical estimate of current capabilities that enables the teacher to generate customized training environments through discrete-time dynamics.

\paragraph{Upper-level Teacher MDP} The teacher operates as an MDP $<S^{u}, A^{u}, P^{u}, R^{u}, \gamma^{u}>$ where:
\begin{itemize}
    \item $S^{u}$: State space representing student performance vector $s^u=p(\pi)=[ p_1,\dots, p_m ]$ across $m$ evaluation environments
    \item $A^{u}$: Action space consisting of environment parameters $\vec{\theta}$ that generate specific environments $\mathcal{M}_{\vec{\theta}}$
    \item $P^{u}$: Transition dynamics reflecting student policy evolution through training
    \item $R^{u}$: Teacher reward (detailed in Section~\ref{sec:reward_design})
\end{itemize}

The teacher observes $s^{u}$, generates environment parameters $a^{u}=\vec{\theta}$, then trains the student for $C$ steps in the resulting environment. Every $C$ steps, the episode ends, and the teacher stores transition $(s^{u}, a^{u}, r^{u}, s^{u,\prime})$ for off-policy training while maximizing cumulative reward within the environment generation budget. Evaluation environment selection is discussed in Section~\ref{sec:reward_evaluation}.

\paragraph{Lower-level Student MDP} The student operates within environments specified by parameters $\vec{\theta}$, forming a POMDP $\mathcal{M}_{\vec{\theta}}=<A,O,S^{\vec{\theta}},\mathcal{P}^{\vec{\theta}},\mathcal{I}^{\vec{\theta}},\mathcal{R}^{\vec{\theta}},\gamma>$ as in standard RL notation~\citep{sutton1998introduction}. The student policy aims to maximize cumulative reward $V^{\vec{\theta}}(\pi) = \sum_{t=0}^C \gamma^t r_t$ over horizon $C$ within each generated environment.

The hierarchical framework enables systematic measurement and adaptation of the student agent's capabilities through the teacher's guided training process. However, collecting student policy evolution trajectories $(s^{u}, a^{u}, r^{u}, s^{u,\prime})$ for teacher training is computationally expensive, as each upper-level MDP transition requires the student to complete a full training horizon of $C$ timesteps in the generated environment. This makes it crucial to minimize the collection of these costly upper-level teacher experiences.

\subsection{Diffusion-Based World Model for Efficient Teacher Training}

SHED employs a conditional diffusion model that functions as a world model~\citep{ha2018world} to simulate student state transitions, significantly reducing the need for costly data collection. This model learns to generate plausible next student states $s^{u,\prime}$ conditioned on current state $s^u$ and teacher action $a^u$.

We formalize this as a conditional diffusion process with:
\begin{equation*}
    q(s_{k}^{u, \prime}|s_{k-1}^{u, \prime}), \quad p_{\phi}(s_{k-1}^{u, \prime}|s_k^{u, \prime},s^{u},a^{u})
\end{equation*}

where $q$ denotes the forward noising process and $p_{\phi}$ represents the trainable reverse denoising process. The model generates synthetic transitions by:

\begin{equation*}
        p_{{\phi}}(s_{0:K}^{u,\prime}|s^{u},a^{u})=\mathcal{N}(s_K^{u,\prime};{0},\mathbf{I}) \prod_{k=1}^K p_{\phi}(s^{u,\prime}_{k-1}|s_k^{u,\prime},s^{u}, a^{u})
\end{equation*}

Following~\citet{ho2020denoising}, we parameterize the reverse process with fixed covariance $\Sigma_{{\phi}}=\beta_i \mathbf{I}$ and mean:
$$\mu_{{\phi}}(s_i^{u,\prime},s^{u}, a^{u},k) =\frac{1}{\sqrt{\alpha_k}}\left( s_k^{u,\prime} - \frac{\beta_k \cdot \epsilon_{\phi}(s_k^{u,\prime},s^{u},a^{u},k)}{\sqrt{1-\bar{\alpha}_k}} \right)$$

These synthetic transitions augment the teacher's experience buffer, enabling more efficient off-policy learning with substantially reduced student interaction requirements.


\textbf{Training Objective.} We employ a similar simplified objective to train the conditional ${\epsilon}$- model:

    \begin{equation}
    \mathcal{L}({\phi}) = 
    \mathbb{E}_{\substack{(s^{u},a^{u},s^{u,\prime} )\sim \tau, \\ k \sim \mathcal{U},{\epsilon} \sim \mathcal{N}({0},\bf{I})}} \left[ \lVert {\epsilon}  - {\epsilon}_{{\phi}}( s^{u,\prime}_k,{s^{u},a^{u}},k)  \rVert^2  \right] 
\end{equation}

where $s^{u,\prime}_k=\sqrt{\bar{\alpha}_k}s^{u,\prime} + \sqrt{1-\bar{\alpha}_k }{\epsilon}$.  The intuition for the loss function $\mathcal{L}({\phi})$ is to predict the noise $\epsilon\sim \mathcal{N}(0,\mathbf{I})$ at the denoising step $k$, and the diffusion model is essentially learning the student policy involution trajectories collected in the real experience buffer $\mathcal{B}_{reals}$. Note that the reverse process necessitates a substantial number of steps $K$~\citep{sohl2015deep}. Recent research by~\citet{xiao2021tackling} has demonstrated that enabling denoising with large steps can reduce the total number of denoising steps $K$. To expedite the relatively slow reverse sampling process (as it requires computing ${\epsilon}_{{\phi}}$ networks $K$ times), we use a small value of $K$. Similar to~\citet{wang2022diffusion}, while simultaneously setting $\beta_{\min} = 0.1$ and $\beta_{\max} = 10.0$, we define:
\begin{equation*}
\begin{aligned}
    \beta_k&=1-\exp\left(\beta_{\min}\times \frac{1}{K} - 0.5(\beta_{\max}-\beta_{\min})\frac{2k -1}{K^2}\right)
\end{aligned}
\end{equation*}
This noise schedule is derived from the variance-preserving Stochastic Differential Equation by~\citet{song2020score}. 

\paragraph{Generate Synthetic Trajectories} 
After training, our diffusion model generates synthetic experience data through a simple process. We start with Gaussian noise $s_K^{u,\prime}\sim \mathcal{N}({0},\mathbf{I})$ and iteratively apply the reverse process $p_{\phi}$ to generate plausible next states $s^{u,\prime}$ conditioned on current state $s^{u}$ and action $a^{u}$.

For actions $a^{u}$, we use random sampling from the action space rather than learning a behavior-cloning diffusion model (as in~\citet{wang2022diffusion}, detailed in the appendix). This straightforward approach increases synthetic experience diversity, helping train a more robust teacher agent.

After generating next state $s^{u,\prime}$, we compute reward $r^{u}$ using the teacher's reward function $R(s^{u}, a^{u},s^{u,\prime})$, creating complete synthetic transitions $(s^{u}, a^{u}, r^{u}, s^{u,\prime})$ for teacher training.

\subsection{Evaluation Environment Selection for Student Policy Representation}\label{sec:reward_evaluation}

The teacher requires accurate assessment of student capabilities to generate appropriate training environments. We represent student policies through performance vectors across carefully selected evaluation environments that comprehensively represent the environment space. Our approach is grounded in the following theoretical result:

\begin{theorem}\label{thm:cover}
There exists a finite evaluation environment set that can capture the student's general capabilities, allowing the performance vector $[p_1, \dots, p_m]$ to serve as an effective representation of student policy $\pi$.
\end{theorem}

The key insight supporting this theorem is that the environment parameter space can be discretized into intervals where student performance remains approximately stable. Under the assumption that small changes in environment parameters result in correspondingly small changes in student performance, we can construct a finite set of ``representative environments'' that comprehensively cover the parameter space $\Theta$. This allows us to represent any student policy through its performance vector across these environments, with a bounded approximation error $\epsilon \rightarrow 0$. For the complete proof, which extends from single-parameter to multi-parameter environment spaces, see Appendix
 \ref{appendix:theorem}.

In practice, our approach leverages domain randomization by first discretizing environment parameters into ranges, then randomly sampling combinations to generate a fixed set of evaluation environments before training begins. This set remains constant throughout the teacher's training period, providing consistent student policy representation.

While Quality-Diversity optimization methods~\citep{fontaine2021differentiable,bhatt2022deep} can generate diverse, high-quality environments, they require substantial computational resources and must be recalculated as student policies evolve. Our straightforward approach proved sufficient in experiments, enabling SHED to surpass baseline algorithms without added complexity. 
 
\subsection{Teacher Reward Design} \label{sec:reward_design}

The teacher's reward function combines two key components: learning progress and fairness in capability development.

Building on Learning Progress concepts \citep{kanitscheider2021multi}, we first measure improvement in student performance across evaluation environments after training:
\begin{equation*}
    R_{progress}(s^u, a^u,s^{u,\prime}) = \sum_{i=1}^{m} (p_i^\prime-p_i)
\end{equation*}

However, optimizing solely for total improvement can lead to uneven development, where gains in some environments come at the cost of deterioration in others. To encourage well-rounded capability development, we incorporate a fairness metric based on the coefficient of variation of performance changes:

\begin{equation}
 cv(s^u, a^u,s^{u,\prime}) = \sqrt{\frac{1}{m-1}\sum_i \frac{(\omega_i-\bar{\omega})^2}{\bar{\omega}^2}}   
\end{equation}

where $\omega_i=p_i^\prime - p_i$ represents the change in student performance for each evaluation environment, and $\bar{\omega}=\frac{1}{m}\sum_{i=1}^m \omega_i$ is the mean performance change. Lower $cv$ values indicate more balanced improvement across environments.

Our final reward function combines both components:
\begin{equation}
    R(s^u, a^u,s^{u,\prime}) = \sum_{i=1}^{m} (p_i^\prime-p_i) - \eta\cdot cv(s^u, a^u,s^{u,\prime})
\end{equation}

where $\eta$ is a coefficient that balances the importance of fairness. This formulation encourages the teacher to generate environments that develop well-rounded student capabilities rather than specialized performance in limited scenarios.

\section{Experiments}\label{sec:exp}
We evaluate SHED against leading approaches under limited teacher-student interactions across three domains: Lunar Lander, Bipedal Walker, and Maze. Our comparison includes:
\begin{itemize}
\item Domain Randomization \citep{tobin2017domain}: Randomly generated environments
\item ACCEL \citep{parker2022evolving}: Evolution-based environment generation that represents the state-of-the-art
\item Edited ACCEL: Modified to exclude revisiting previously generated environments
\item PAIRED \citep{dennis2020emergent}: Adversarial environment generation
\item h-MDP: Our hierarchical approach without the diffusion component (ablation)
\end{itemize}

For each domain, we construct two distinct environment sets. First, evaluation environments following the method described in Section \ref{sec:reward_evaluation} are used by SHED's teacher to assess student capabilities. Second, test environments are generated randomly but fixed across all algorithms to enable fair comparison. To ensure accurate measurement of zero-shot performance, we carefully exclude both evaluation and test environments from the training process, maintaining them as completely separate sets.

\paragraph{Resource-Constrained Experimental Design} We implement an experimental setup that departs from standard UED approaches by strictly limiting teacher-student interactions. Rather than allowing unlimited environment generation, we constrain each teacher to a fixed budget of environments per episode (50 in our implementation). After reaching this interaction limit, the student agent is reset and a new episode begins, allowing us to evaluate the teacher's ability to efficiently guide policy development from scratch within tight resource constraints.

This setup specifically simulates real-world deployment scenarios where interaction costs or resource limitations necessitate efficiency. Unlike traditional UED methods that require continuous environment generation throughout training, our goal through this experiments is to demonstrate how our method can develop capable student agents using only 50 environments during deployment. This efficiency is particularly valuable in practical applications where environmental interactions are costly, time-consuming, or otherwise limited.

\subsection{Setup} 
\begin{figure*}[ht]
    \centering
    \begin{minipage}[b]{0.49\linewidth}
        \includegraphics[width=\linewidth]{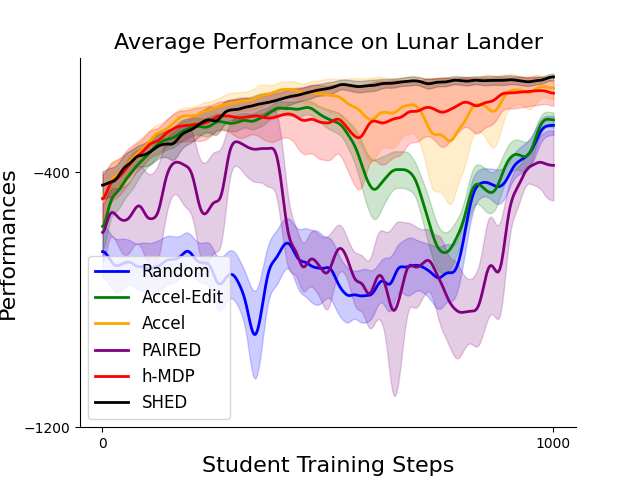}
    \end{minipage}
    \hfill
    \begin{minipage}[b]{0.49\linewidth}
        \includegraphics[width=\linewidth]{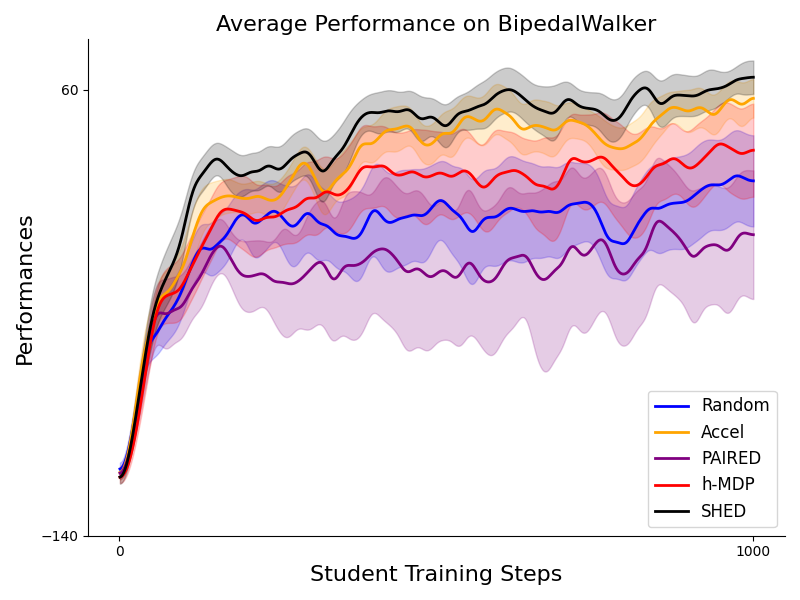}
    \end{minipage}
    \caption{Mean zero-shot transfer performance on test environments during the final teacher episode for Lunar Lander (left) and BipedalWalker (right). Shading indicates standard error across five independent runs with different random seeds.}
    \label{fig:performance}
\end{figure*}

We implemented our experiments with consistent algorithms and protocols across all conditions. Student agents were trained using Proximal Policy Optimization (PPO; \citet{schulman2017proximal}), while teacher agents in SHED and h-MDP employed Deterministic Policy Gradient algorithms (DDPG; \citet{silver2014deterministic}), selected for their ability to learn from both real and synthetic experiences in an off-policy manner.

Our experimental protocol consisted of 50 teacher episodes, with each teacher generating 50 environments per episode. Within each environment, a student PPO agent completed four training epochs using five PPO mini-batches, resulting in 1000 PPO updates per teacher episode. Each episode concluded with the initialization of a new student for the subsequent episode.

In this episodic framework, ACCEL maintained its replay buffer across episodes, while ACCEL-Edit reset the buffer at the start of each teacher episode. This setup differs from the original ACCEL implementation, which employed 30,000 PPO updates throughout the entire experimental run \citep{parker2022evolving}.

The experiments span three domains with varying complexity:

For Lunar Lander, students must safely land a vehicle while teachers control environmental parameters including gravity, wind power, and turbulence. Student training runs for 1 million environment steps.

In BipedalWalker, adapted from \citet{parker2022evolving}, students guide a bipedal walker across 2D terrain. Teachers generate environments by adjusting eight parameters controlling obstacle features like stump height and ground roughness. Student training extends to 10 million steps.

The Maze domain tasks students with finding a goal while limited to a 3x3 grid field of vision. Rather than using a high-dimensional parameter space to specify maze layouts directly, we use ChatGPT to generate feasible mazes based on high-level feature descriptions (size, structure). Teachers specify these abstracted parameters as discrete values. Student training runs for 400,000 steps, with details on maze generation in Appendix~\ref{app:ssub:maze_doc} and prompts in Appendix~\ref{app:ssub:prompt}.

For SHED and h-MDP, we employ 10, 15, and 10 evaluation environments as teacher observations for LunarLander, BipedalWalker, and Maze respectively, with an ablation study on observation size provided in Appendix \ref{app:ssec:ablation}. Additional experimental details and hyperparameters are available in the appendix.

\section{Results and Discussion}
\begin{wrapfigure}{l}{0.45\textwidth}
 \centering
 \includegraphics[width=\linewidth]{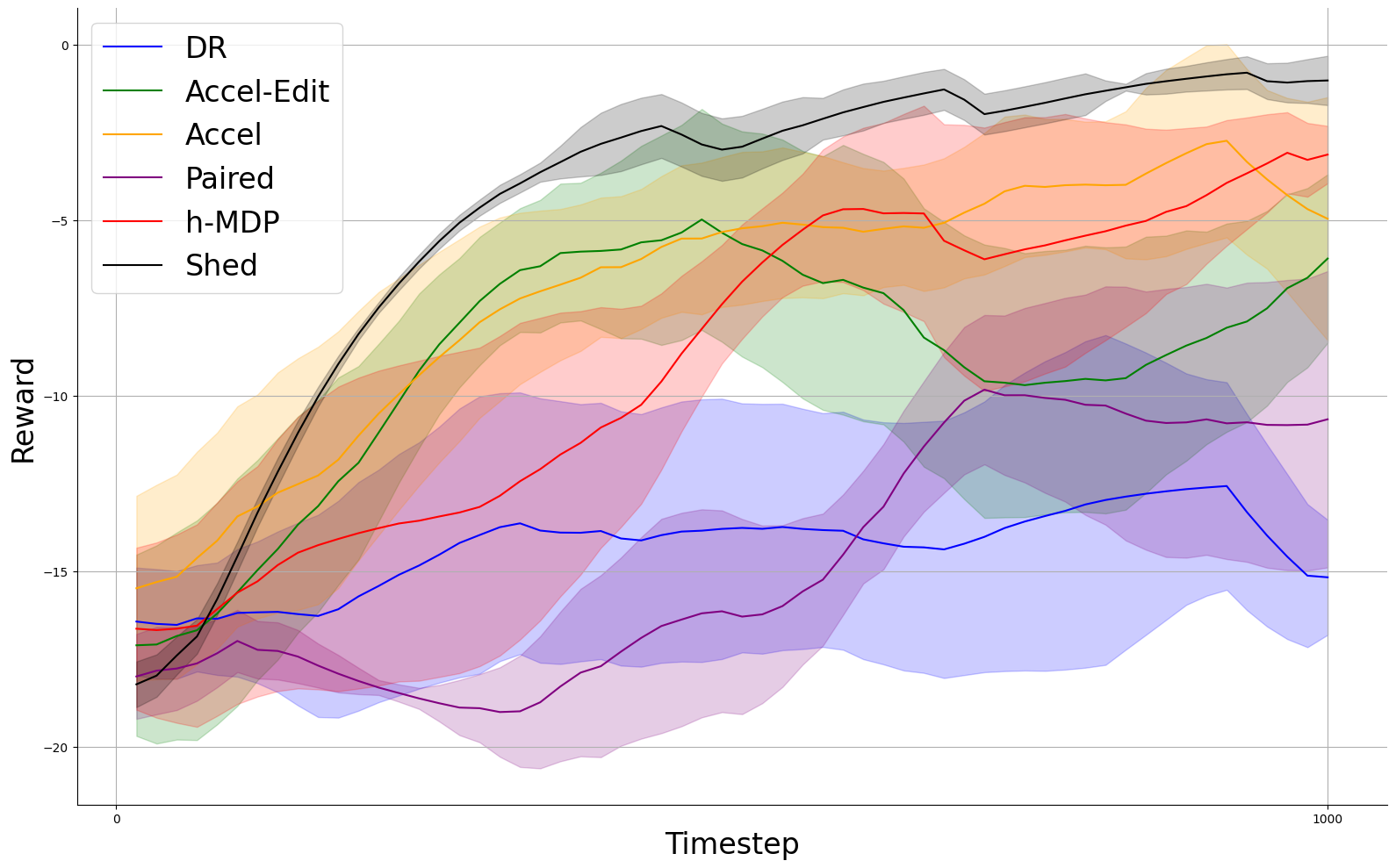}
 \caption{Mean rewards on Maze test environments during the final teacher episode. Shading indicates standard error across five independent runs.}
 \label{fig:maze_results}
 \vspace{-25pt}
\end{wrapfigure}
Our experiments demonstrate SHED's superior performance across all tested domains. Students trained with SHED consistently achieved higher zero-shot performance on unseen test environments compared to all baseline methods, including ACCEL, despite operating under strict resource constraints.

\begin{wrapfigure}{r}{0.45\textwidth}
 \vspace{-5pt}
 \centering
 \includegraphics[width=\linewidth]{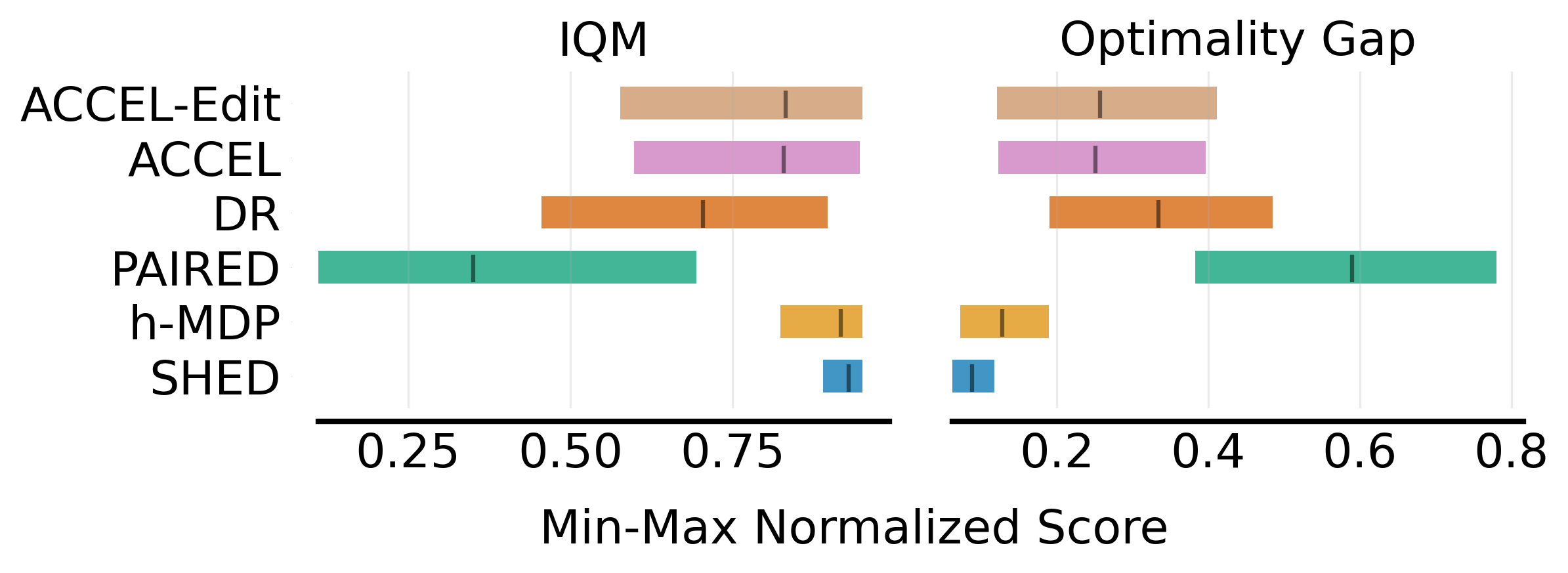}
 \vspace{-10pt}
 \caption{Normalized performance over 5 runs in Maze domain. Higher IQM scores and lower optimality gaps indicate high performance.}
  \vspace{-5pt}
 \label{fig:maze_results_iqm}
\end{wrapfigure}
Figures \ref{fig:performance} and \ref{fig:maze_results} present the average performance on test environments during the final teacher episode for all three domains, across five seeds. SHED exhibited exceptional consistency between runs, as evidenced by the narrow standard error bands and high Interquartile Mean (IQM; \citet{agarwal2021deep}; Figure \ref{fig:maze_results_iqm}) with minimal variation. This stability contrasts sharply with baseline approaches, particularly PAIRED, which showed substantial performance fluctuations and wide IQM ranges across different seeds. Furthermore, SHED achieved the lowest Optimality Gap scores, indicating its effectiveness in training student agents that generalize well to unseen environments.

During Lunar Lander training, we observed that baseline algorithms experienced a notable performance decline midway through training, suggesting difficulty in adapting to students' evolving capabilities. In contrast, both SHED and h-MDP maintained stable performance throughout the training process by actively predicting student progression rather than merely reacting based on student history. SHED demonstrated this adaptability most effectively, with h-MDP showing similar but less pronounced benefits.
The performance gap between SHED and h-MDP highlights the significant benefits of our diffusion-based synthetic data generation approach. By augmenting teacher training with synthetic experiences, SHED enables more effective environment selection tailored to student capabilities.

The performance disparity between ACCEL and ACCEL-Edit—where ACCEL-Edit clears its replay buffer after each episode—emphasizes the critical importance of preserving valuable past environments. Our h-MDP approach matched ACCEL's performance in Lunar Lander and Maze domains while exceeding ACCEL-Edit, highlighting a key strength: the ability to learn from previous environments implicitly through reinforcement learning, similar to ACCEL's explicit storage mechanism but with greater flexibility.

Finally, SHED demonstrates remarkably consistent performance across all experiments, as evidenced by the minimal standard error (narrow shaded regions) compared to baseline methods. This consistency likely stems from our careful selection of evaluation environments through parameter discretization. In our ablation study (Appendix \ref{app:subsec:div_eval_envs}), we found that while purely random environment selection without discretization can sometimes achieve similar or even superior performance, it risks including out-of-distribution (OOD) environments that are infeasible or unsolvable, potentially leading to poorer performance.

\paragraph{Ablation and Additional Experiments} Additional ablation studies are detailed in the Appendix, including: diffusion model effectiveness in generating synthetic student policy trajectories (Appendix~\ref{app:sec:diffusion}); impact of evaluation environment set size and training horizon (Appendix~\ref{app:ssec:ablation}); and behavior under increased budget constraints and higher fairness reward weights (Appendix~\ref{app:ssec:lunar_lander_add}). Overall, the results indicate that our design choices were ideal.


\section{Conclusion and Limitations}
We introduced SHED, an adaptive framework for training generally capable agents with a limited budget. Our approach formalizes teacher-student interaction as a hierarchical MDP, with a diffusion-based generative model that creates synthetic training data for the teacher. Across multiple environments, SHED consistently outperformed baseline algorithms in producing agents that perform in unseen conditions.
While SHED requires an initial investment in teacher training (approximately 2500 environments across 50 episodes), it yields a specialized teacher policy that can efficiently train new student agents using only 50 environments. This efficiency is particularly valuable when training multiple students or when teacher-student interactions are costly, as the initial investment is amortized across subsequent training sessions.

SHED offers a promising alternative for sim-to-real challenge: rather than transferring pre-trained policies through domain randomization \citep{duan2024learningvisionbasedbipedallocomotion,yuan2024learningmanipulateanywherevisual}, we transfer a teacher that efficiently adapts new policies with minimal real-world interaction where data collection is costly. This approach assumes learning dynamics are comparable between simulated and real environments, an assumption requiring further investigation.

A key limitation is that SHED's RL-based method may lose efficiency in scenarios with high-dimensional teacher action spaces, where domain randomization approaches may scale better. This limitation warrants further investigation in future work.
\clearpage
\bibliography{neurips25/neurips_2025}

\begin{thebibliography}{32}
\providecommand{\natexlab}[1]{#1}
\providecommand{\url}[1]{\texttt{#1}}
\expandafter\ifx\csname urlstyle\endcsname\relax
  \providecommand{\doi}[1]{doi: #1}\else
  \providecommand{\doi}{doi: \begingroup \urlstyle{rm}\Url}\fi

\bibitem[Agarwal et~al.(2021)Agarwal, Schwarzer, Castro, Courville, and Bellemare]{agarwal2021deep}
Rishabh Agarwal, Max Schwarzer, Pablo~Samuel Castro, Aaron~C Courville, and Marc Bellemare.
\newblock Deep reinforcement learning at the edge of the statistical precipice.
\newblock \emph{Advances in neural information processing systems}, 34:\penalty0 29304--29320, 2021.

\bibitem[Azad et~al.(2023)Azad, Gur, Emhoff, Alexis, Faust, Abbeel, and Stoica]{azad2023clutr}
Abdus~Salam Azad, Izzeddin Gur, Jasper Emhoff, Nathaniel Alexis, Aleksandra Faust, Pieter Abbeel, and Ion Stoica.
\newblock Clutr: Curriculum learning via unsupervised task representation learning.
\newblock In \emph{International Conference on Machine Learning}, pages 1361--1395. PMLR, 2023.

\bibitem[Bhatt et~al.(2022)Bhatt, Tjanaka, Fontaine, and Nikolaidis]{bhatt2022deep}
Varun Bhatt, Bryon Tjanaka, Matthew Fontaine, and Stefanos Nikolaidis.
\newblock Deep surrogate assisted generation of environments.
\newblock \emph{Advances in Neural Information Processing Systems}, 35:\penalty0 37762--37777, 2022.

\bibitem[Costales and Nikolaidis(2024)]{costales2024enabling}
Robby Costales and Stefanos Nikolaidis.
\newblock Enabling adaptive agent training in open-ended simulators by targeting diversity.
\newblock \emph{arXiv preprint arXiv:2411.04466}, 2024.

\bibitem[Da et~al.(2025)Da, Turnau, Kutralingam, Velasquez, Shakarian, and Wei]{da2025survey}
Longchao Da, Justin Turnau, Thirulogasankar~Pranav Kutralingam, Alvaro Velasquez, Paulo Shakarian, and Hua Wei.
\newblock A survey of sim-to-real methods in rl: Progress, prospects and challenges with foundation models.
\newblock \emph{arXiv preprint arXiv:2502.13187}, 2025.

\bibitem[Dennis et~al.(2020)Dennis, Jaques, Vinitsky, Bayen, Russell, Critch, and Levine]{dennis2020emergent}
Michael Dennis, Natasha Jaques, Eugene Vinitsky, Alexandre Bayen, Stuart Russell, Andrew Critch, and Sergey Levine.
\newblock Emergent complexity and zero-shot transfer via unsupervised environment design.
\newblock \emph{Advances in neural information processing systems}, 33:\penalty0 13049--13061, 2020.

\bibitem[Duan et~al.(2024)Duan, Pandit, Gadde, van Marum, Dao, Kim, and Fern]{duan2024learningvisionbasedbipedallocomotion}
Helei Duan, Bikram Pandit, Mohitvishnu~S. Gadde, Bart van Marum, Jeremy Dao, Chanho Kim, and Alan Fern.
\newblock Learning vision-based bipedal locomotion for challenging terrain, 2024.
\newblock URL \url{https://arxiv.org/abs/2309.14594}.

\bibitem[Fontaine and Nikolaidis(2021)]{fontaine2021differentiable}
Matthew Fontaine and Stefanos Nikolaidis.
\newblock Differentiable quality diversity.
\newblock \emph{Advances in Neural Information Processing Systems}, 34:\penalty0 10040--10052, 2021.

\bibitem[Garcin et~al.(2024)Garcin, Doran, Guo, Lucas, and Albrecht]{garcin2024dred}
Samuel Garcin, James Doran, Shangmin Guo, Christopher~G Lucas, and Stefano~V Albrecht.
\newblock Dred: Zero-shot transfer in reinforcement learning via data-regularised environment design.
\newblock \emph{arXiv preprint arXiv:2402.03479}, 2024.

\bibitem[Ha and Schmidhuber(2018)]{ha2018world}
David Ha and J{\"u}rgen Schmidhuber.
\newblock World models.
\newblock \emph{arXiv preprint arXiv:1803.10122}, 2018.

\bibitem[Ho et~al.(2020)Ho, Jain, and Abbeel]{ho2020denoising}
Jonathan Ho, Ajay Jain, and Pieter Abbeel.
\newblock Denoising diffusion probabilistic models.
\newblock \emph{Advances in neural information processing systems}, 33:\penalty0 6840--6851, 2020.

\bibitem[Jiang et~al.(2021)Jiang, Grefenstette, and Rockt{\"a}schel]{jiang2021prioritized}
Minqi Jiang, Edward Grefenstette, and Tim Rockt{\"a}schel.
\newblock Prioritized level replay.
\newblock In \emph{International Conference on Machine Learning}, pages 4940--4950. PMLR, 2021.

\bibitem[Kanitscheider et~al.(2021)Kanitscheider, Huizinga, Farhi, Guss, Houghton, Sampedro, Zhokhov, Baker, Ecoffet, Tang, et~al.]{kanitscheider2021multi}
Ingmar Kanitscheider, Joost Huizinga, David Farhi, William~Hebgen Guss, Brandon Houghton, Raul Sampedro, Peter Zhokhov, Bowen Baker, Adrien Ecoffet, Jie Tang, et~al.
\newblock Multi-task curriculum learning in a complex, visual, hard-exploration domain: Minecraft.
\newblock \emph{arXiv preprint arXiv:2106.14876}, 2021.

\bibitem[Kingma(2013)]{kingma2013auto}
Diederik~P Kingma.
\newblock Auto-encoding variational bayes.
\newblock \emph{arXiv preprint arXiv:1312.6114}, 2013.

\bibitem[Li et~al.(2023)Li, Li, and Varakantham]{li2023diversity}
Dexun Li, Wenjun Li, and Pradeep Varakantham.
\newblock Diversity induced environment design via self-play.
\newblock \emph{arXiv preprint arXiv:2302.02119}, 2023.

\bibitem[Li et~al.(2025)Li, Li, and Varakantham]{Li_Li_Varakantham_2025}
Dexun Li, Wenjun Li, and Pradeep Varakantham.
\newblock Marginal benefit driven rl teacher for unsupervised environment design.
\newblock \emph{Proceedings of the AAAI Conference on Artificial Intelligence}, 39\penalty0 (17):\penalty0 18253--18261, Apr. 2025.
\newblock \doi{10.1609/aaai.v39i17.34008}.
\newblock URL \url{https://ojs.aaai.org/index.php/AAAI/article/view/34008}.

\bibitem[Nichol et~al.(2021)Nichol, Dhariwal, Ramesh, Shyam, Mishkin, McGrew, Sutskever, and Chen]{nichol2021glide}
Alex Nichol, Prafulla Dhariwal, Aditya Ramesh, Pranav Shyam, Pamela Mishkin, Bob McGrew, Ilya Sutskever, and Mark Chen.
\newblock Glide: Towards photorealistic image generation and editing with text-guided diffusion models.
\newblock \emph{arXiv preprint arXiv:2112.10741}, 2021.

\bibitem[Nie et~al.(2022)Nie, Guo, Huang, Xiao, Vahdat, and Anandkumar]{nie2022diffusion}
Weili Nie, Brandon Guo, Yujia Huang, Chaowei Xiao, Arash Vahdat, and Anima Anandkumar.
\newblock Diffusion models for adversarial purification.
\newblock \emph{arXiv preprint arXiv:2205.07460}, 2022.

\bibitem[Parker-Holder et~al.(2022)Parker-Holder, Jiang, Dennis, Samvelyan, Foerster, Grefenstette, and Rockt{\"a}schel]{parker2022evolving}
Jack Parker-Holder, Minqi Jiang, Michael Dennis, Mikayel Samvelyan, Jakob Foerster, Edward Grefenstette, and Tim Rockt{\"a}schel.
\newblock Evolving curricula with regret-based environment design.
\newblock \emph{arXiv preprint arXiv:2203.01302}, 2022.

\bibitem[Saharia et~al.(2022)Saharia, Chan, Saxena, Li, Whang, Denton, Ghasemipour, Gontijo~Lopes, Karagol~Ayan, Salimans, et~al.]{saharia2022photorealistic}
Chitwan Saharia, William Chan, Saurabh Saxena, Lala Li, Jay Whang, Emily~L Denton, Kamyar Ghasemipour, Raphael Gontijo~Lopes, Burcu Karagol~Ayan, Tim Salimans, et~al.
\newblock Photorealistic text-to-image diffusion models with deep language understanding.
\newblock \emph{Advances in Neural Information Processing Systems}, 35:\penalty0 36479--36494, 2022.

\bibitem[Schulman et~al.(2017)Schulman, Wolski, Dhariwal, Radford, and Klimov]{schulman2017proximal}
John Schulman, Filip Wolski, Prafulla Dhariwal, Alec Radford, and Oleg Klimov.
\newblock Proximal policy optimization algorithms.
\newblock \emph{arXiv preprint arXiv:1707.06347}, 2017.

\bibitem[Silver et~al.(2014)Silver, Lever, Heess, Degris, Wierstra, and Riedmiller]{silver2014deterministic}
David Silver, Guy Lever, Nicolas Heess, Thomas Degris, Daan Wierstra, and Martin Riedmiller.
\newblock Deterministic policy gradient algorithms.
\newblock In \emph{International conference on machine learning}, pages 387--395. Pmlr, 2014.

\bibitem[Sohl-Dickstein et~al.(2015)Sohl-Dickstein, Weiss, Maheswaranathan, and Ganguli]{sohl2015deep}
Jascha Sohl-Dickstein, Eric Weiss, Niru Maheswaranathan, and Surya Ganguli.
\newblock Deep unsupervised learning using nonequilibrium thermodynamics.
\newblock In \emph{International conference on machine learning}, pages 2256--2265. PMLR, 2015.

\bibitem[Song et~al.(2020)Song, Sohl-Dickstein, Kingma, Kumar, Ermon, and Poole]{song2020score}
Yang Song, Jascha Sohl-Dickstein, Diederik~P Kingma, Abhishek Kumar, Stefano Ermon, and Ben Poole.
\newblock Score-based generative modeling through stochastic differential equations.
\newblock \emph{arXiv preprint arXiv:2011.13456}, 2020.

\bibitem[Sutton et~al.(1998)Sutton, Barto, et~al.]{sutton1998introduction}
Richard~S Sutton, Andrew~G Barto, et~al.
\newblock \emph{Introduction to reinforcement learning}, volume 135.
\newblock MIT press Cambridge, 1998.

\bibitem[Tashiro et~al.(2021)Tashiro, Song, Song, and Ermon]{tashiro2021csdi}
Yusuke Tashiro, Jiaming Song, Yang Song, and Stefano Ermon.
\newblock Csdi: Conditional score-based diffusion models for probabilistic time series imputation.
\newblock \emph{Advances in Neural Information Processing Systems}, 34:\penalty0 24804--24816, 2021.

\bibitem[Tio and Varakantham(2023)]{tio2023transferable}
Sidney Tio and Pradeep Varakantham.
\newblock Transferable curricula through difficulty conditioned generators.
\newblock \emph{arXiv preprint arXiv:2306.13028}, 2023.

\bibitem[Tobin et~al.(2017)Tobin, Fong, Ray, Schneider, Zaremba, and Abbeel]{tobin2017domain}
Josh Tobin, Rachel Fong, Alex Ray, Jonas Schneider, Wojciech Zaremba, and Pieter Abbeel.
\newblock Domain randomization for transferring deep neural networks from simulation to the real world.
\newblock In \emph{2017 IEEE/RSJ international conference on intelligent robots and systems (IROS)}, pages 23--30. IEEE, 2017.

\bibitem[Wang et~al.(2023)Wang, Hunt, and Zhou]{wang2022diffusion}
Zhendong Wang, Jonathan~J Hunt, and Mingyuan Zhou.
\newblock Diffusion policies as an expressive policy class for offline reinforcement learning.
\newblock In \emph{The Eleventh International Conference on Learning Representations}, 2023.
\newblock URL \url{https://openreview.net/forum?id=AHvFDPi-FA}.

\bibitem[Wyatt et~al.(2022)Wyatt, Leach, Schmon, and Willcocks]{wyatt2022anoddpm}
Julian Wyatt, Adam Leach, Sebastian~M Schmon, and Chris~G Willcocks.
\newblock Anoddpm: Anomaly detection with denoising diffusion probabilistic models using simplex noise.
\newblock In \emph{Proceedings of the IEEE/CVF Conference on Computer Vision and Pattern Recognition}, pages 650--656, 2022.

\bibitem[Xiao et~al.(2021)Xiao, Kreis, and Vahdat]{xiao2021tackling}
Zhisheng Xiao, Karsten Kreis, and Arash Vahdat.
\newblock Tackling the generative learning trilemma with denoising diffusion gans.
\newblock \emph{arXiv preprint arXiv:2112.07804}, 2021.

\bibitem[Yuan et~al.(2024)Yuan, Wei, Cheng, Zhang, Chen, and Xu]{yuan2024learningmanipulateanywherevisual}
Zhecheng Yuan, Tianming Wei, Shuiqi Cheng, Gu~Zhang, Yuanpei Chen, and Huazhe Xu.
\newblock Learning to manipulate anywhere: A visual generalizable framework for reinforcement learning, 2024.
\newblock URL \url{https://arxiv.org/abs/2407.15815}.

\end{thebibliography}
\bibliographystyle{plainnat}
\clearpage
\appendix
\section*{Technical Appendices and Supplementary Material}
\section{Preliminaries for Diffusion Modeling}
\label{app:preliminary}
Diffusion models~\citep{sohl2015deep} are a specific type of generative model that learns the data distribution. Recent advances in diffusion-based models, including Langevin dynamics and score-based generative models, have shown promising results in various applications, such as time series forecasting~\citep{tashiro2021csdi}, robust learning~\citep{nie2022diffusion}, anomaly detection~\citep{wyatt2022anoddpm} as well as synthesizing high-quality images from text descriptions~\citep{nichol2021glide,saharia2022photorealistic}. These models can be trained using standard optimization techniques, such as stochastic gradient descent, making them highly scalable and easy to implement.

In a diffusion probabilistic model, we assume a $d$-dimensional random variable $x_0 \in \mathbb{R}^d$ with an unknown distribution $q(x_0)$. Diffusion Probabilistic model involves two Markov chains: a predefined forward chain $\displaystyle q(x_k|x_{k-1})$ that perturbs data to noise, and a trainable reverse chain $\displaystyle p_{\phi}(x_{k-1}|x_{k})$ that converts noise back to data. The forward chain is typically designed to transform any data distribution into a simple prior distribution (e.g., standard Gaussian) by considering perturb data with Gaussian noise of zero mean and a fixed variance schedule $\{ \beta_k\}_{k=1}^K$ for $K$ steps:
\begin{equation}
\begin{aligned}
        q(x_k|x_{k-1}) &=  \mathcal{N} ( x_k ; \sqrt{1-\beta_k}x_{k-1} , \beta_t \mathbf{I}) \quad \text{and} \\
         q(x_{1:K}|x_{0}) &= \Pi_{k=1}^K q(x_k|x_{k-1}),
\end{aligned}
\end{equation}\label{eq:forward_chain}
\noindent where $k\in\{ 1,\dots, K\}$, and $0<\beta_{1:K}<1$ denote the noise scale scheduling. As $K \rightarrow \infty$, $x_K$ will converge to isometric Gaussian noise: $x_K \rightarrow \mathcal{N} (0,\bf{I})$. According to the rule of the sum of normally distributed random variables, the choice of Gaussian noise provides a closed-form solution to generate arbitrary time-step $x_k$ through:
\begin{equation}\label{eq:reverse_chain}
    x_k = \sqrt{\bar{\alpha}_k}x_0 + \sqrt{1-\bar{\alpha}_k }{\epsilon}, \quad \text{where} \quad {\epsilon}\sim \mathcal{N} (0,\bf{I}).
\end{equation}
\noindent Here $\alpha_k = 1- \beta_k$ and $\bar{\alpha}_k = \prod_{s=1}^k \alpha_s$. The reverse chain
$p_{\phi}(x_{k-1}|x_k)$ reverses the forward process by learning transition kernels parameterized by deep neural networks. Specifically, considering the Markov chain parameterized by $\phi$, denoising arbitrary Gaussian noise into clean data samples can be written as:
\begin{equation}\label{eq:reverse_chain_related}
    p_{\phi}(x_{k-1}|x_k) =  \mathcal{N} ( x_{k-1} ; {\mu}_{\phi}(x_k,k) , \Sigma_{{\phi}}(x_k,k))
\end{equation}
It uses the Gaussian form $ p_{\phi}(x_{k-1}|x_k)$ because the reverse process has the identical function form as the forward process when $\beta_t$ is small~\citep{sohl2015deep}. \citet{ho2020denoising} consider the following parameterization of $ p_{\phi}(x_{k-1}|x_k)$:
\begin{equation}
    \begin{aligned}
        {\mu}_{\phi}(x_k,k) &= \frac{1}{\alpha_k}\left( x_k - \frac{\beta_k}{\sqrt{1-\alpha_k}} \epsilon_{\phi}(x_k,k) \right) \text{ and } \\
        \Sigma_{{\phi}}(x_k,k)&=\tilde{\beta}_k^{1/2} \text{ where } \tilde{\beta}_k=\begin{cases} 
\frac{1-\alpha_{k-1}}{1-\alpha_k} \beta_k & k>1 \\
\beta_1 & k=1
\end{cases}
    \end{aligned}
\end{equation}\label{eq:parameterization}
\noindent $\bf{\epsilon}_{\phi}$ is a trainable function to predict the noise vector $\bf{\epsilon}$ from $x_k$. 
\citet{ho2020denoising} show that training the reverse chain to maximize the log-likelihood $\int q(x_0) \log p_{\phi}(x_{0}) d x_{0}$ is equivalent to minimizing re-weighted evidence lower bound (ELBO) that fits the noise. They derive the final simplified optimization objective:
\begin{equation}
    \mathcal{L}(\phi) = \mathbb{E}_{x_0,k,{\epsilon}} \left[ \lVert {\epsilon}  - {\epsilon}_{\phi}(\sqrt{\bar{\alpha}_k}x_0 + \sqrt{1-\bar{\alpha}_k }{\epsilon}, k)  \rVert^2  \right]
\end{equation}\label{eq:objective}
\noindent Once the model is trained, new data points can be subsequently generated by first sampling a random vector from the prior distribution, followed by ancestral sampling through the reverse Markov chain in Equation~\ref{eq:reverse_chain_related}. 

\section{Theorem}
\label{appendix:theorem}
\begin{theorem}\label{thm:cover2}
There exists a finite evaluation environment set that can capture the student's general capabilities and the performance vector $[p_1, \dots, p_m]$ is a good representation of the student policy. 
\end{theorem}

To prove this, we first provide the following Assumption:
\begin{assumption}
    Let $p(\pi,{\vec{\theta}})$ denote the performance of student policy $\pi$ in an environment $\vec{\theta}$. For $\forall i$-th dimension of the environment parameters, denoted as $\theta_i$, when changing the $\theta_i$ to $\theta_i^\prime$ to get a new environment $\vec{\theta^\prime}$ while keeping other environment parameters fixed, there $\exists \delta_i>0$,  if $|\theta_i^\prime-\theta_i|\leq \delta_i$, we have $|p(\pi, \vec{\theta^\prime}) - p(\pi, \vec{\theta})|\leq \epsilon_i$, where $\epsilon_i \rightarrow 0$.
\end{assumption}
If this is true, we then can construct a finite set of environments, and the student performances in those environments can represent the performances in all potential environments generated within the certain environment parameters open interval combinations, and the set of those open intervals combinations cover the environment parameter space $\Theta$.

We begin from the simplest case where we only consider using one environment parameter to generate environments, denoted as $\theta_i$. We can construct a finite environment parameter set for environment parameters, which is $\{ \theta_i^{min}+1/2*\delta_i, \theta_i^{min}+3/2*\delta_i, \theta_i^{min}+7/2*\delta_i,\dots, \theta_i^{max}-\delta_i/2 \}$. Assume the set size is $L_i$. We let the set $\{\vec{\theta_i} \}_{i=1}^{L_i}$ denote the corresponding generated environments. This is served as the \textbf{representative environment set}. Then the student performances in those environments are denoted as $\{p(\pi, \vec{\theta_i})\}_{i=1}^{L_i}$, which we call it as \textbf{representative performance vector set}. We can divide the space for $\theta_i$ into a finite set of open intervals with size $L_i$, which is 
 $\{ [\theta_i^{min}, \theta_i^{min}+3/2*\delta_i), (\theta_i^{min}+1/2*\delta_i,\theta_i^{min}+5/2\delta_i), (\theta_i^{min}+5/2*\delta_i,\theta_i^{min}+9/2*\delta_i),\dots, (\theta_i^{max}-3/2*\delta_i,\theta_i^{max}] \}$, which we call it as \textbf{representative parameter interval set}, also denoted as $\{(\theta_i-\delta, \theta_i+\delta)\}_{i=1}^{L_i}$. For any environment generated in those intervals, denoted as $\vec{\theta_i^\prime}$, the performance $p(\pi, \vec{\theta_i^\prime})$ can always be represented by the $p(\pi, \vec{\theta_i})$ which is in the same interval, as $|p(\pi, \vec{\theta^\prime_i}) - p(\pi, \vec{\theta_i})|\leq \epsilon_i$, where $\epsilon_i \rightarrow 0$. In such cases, the finite set of environmental parameter intervals $\{ \theta_i^{min}+1/2*\delta_i, \theta_i^{min}+3/2*\delta_i, \theta_i^{min}+7/2*\delta_i,\dots, \theta_i^{max}-\delta_i/2 \}$ fully covers the entire parameter space $\Theta$. We can find a {representative environment set}  $\{\vec{\theta_i} \}_{i=1}^{L_i}$ that is capable of approximating the performance of the student policy within the open parameter intervals combination. This set effectively characterizes the general performance capabilities of the student policy $\pi$.

Then we extend to two environment parameter design space cases. Let's assume that the environment is generated by two-dimension environment parameters. Then, for each environment parameter, $\theta_i \in \{ \theta_1, \theta_2\}$. We can find the same open interval set for each parameter. Specifically, for each $\theta_i$, there exists a $\delta_i$, such that  if $|\theta_i^\prime-\theta_i|\leq \delta_i$, we have $|p(\pi, \vec{\theta^\prime}) - p(\pi, \vec{\theta})|\leq \epsilon_i$, where $\epsilon_i \rightarrow 0$. Hence, we let $\delta = \min \{ \delta_1, \delta_2\}$ and $\epsilon = \epsilon_1 + \epsilon_2$. Thus the new \textbf{representative environment set} is the set that includes the any combination of $\{ [\theta_1, \theta_2] \}$ where $\theta_1\in \{\vec{\theta_i} \}_{i=1}^{L_1}$ and $\theta_2 \in \{\vec{\theta_j} \}_{j=1}^{L_2}$. We can get the \textbf{representative performance vector set} as $\{p(\pi, [\vec{\theta_i},\vec{\theta_j}])\}_{i\in [1,L_1],j\in [1,L_2]}$. We then can construct the \textbf{representative parameter interval set} as  $\{[(\theta_i-\delta, \theta_i+\delta), (\theta_j-\delta, \theta_j+\delta)]\}_{i\in [1,L_1], j\in [1,L_j]}$. As a result, for any new environments $[\vec{\theta_i^\prime},\vec{\theta_j^\prime}]$, we can find the representative environment whose environment parameters are in the same parameter interval $[\vec{\theta_i},\vec{\theta_j}]$, such that their performance difference is smaller than $\epsilon = \epsilon_1 + \epsilon_2$ for all $\forall i\in[1,L_1] ,\forall j\in[1,L_2]$:
\begin{equation}
   \begin{aligned}
       &|p(\pi, [\vec{\theta_i^\prime},\vec{\theta_j^\prime}])- p(\pi, [\vec{\theta_i},\vec{\theta_j}])| = \\
       &|p(\pi, [\vec{\theta_i^\prime},\vec{\theta_j^\prime}]) -p(\pi, [\vec{\theta_i^\prime},\vec{\theta_j}]) + p(\pi, [\vec{\theta_i^\prime},\vec{\theta_j}])- p(\pi, [\vec{\theta_i},\vec{\theta_j}])| \\
       &\leq |p(\pi, [\vec{\theta_i^\prime},\vec{\theta_j^\prime}]) -p(\pi, [\vec{\theta_i^\prime},\vec{\theta_j}])| + \\
       &\quad|p(\pi, [\vec{\theta_i^\prime},\vec{\theta_j}])- p(\pi, [\vec{\theta_i},\vec{\theta_j}])| \\
       &\leq \delta_j+\delta_i = \delta        
   \end{aligned}
\end{equation}

 In such cases, the finite set of environmental parameter intervals $\{[(\theta_i-\delta, \theta_i+\delta), (\theta_j-\delta, \theta_j+\delta)]\}_{i\in [1,L_1], j\in [1,L_j]}$ fully covers the entire parameter space $\Theta$. We can find a {representative environment set}  $\{\vec{\theta_i} \}_{i=1}^{L_i}$ that is capable of approximating the performance of the student policy within the open parameter intervals combination. This set effectively characterizes the general performance capabilities of the student policy $\pi$.

Similarly, we can show this still holds when the environment is constructed by a larger dimension environment parameters,  where we set $\delta = \min \{ \delta_i \}$, and $\epsilon=\sum_i \epsilon_i$, and we have $\delta>0$, $\epsilon\rightarrow 0$. The overall logic is that we can find a finite set, which is called \textbf{representative environment set}, and we can use performances in this set to represent any performances in the environments generated in the \textbf{representative parameter interval set}, which is called  \textbf{representative performance vector set}. Finally, we can show that \textbf{representative parameter interval set} fully covers the environment parameter space. Thus there exists a finite evaluation environment set that can capture the student's general capabilities and the performance vector, called \textbf{representative performance vector set}, $[p_1, \dots, p_m]$ is a good representation of the student policy.

\begin{table*}[t]
\centering

\begin{tabular}{lll}
\hline
UED Approaches    & Teacher Policy & Decision Rule \\ \midrule
{DR}~\cite{tobin2017domain}      & $\Lambda(\pi)=U(\Theta) $           & Randomly sample \\ 
{PAIRED}~\citep{dennis2020emergent}  & $\Lambda(\pi)=\{ \bar{\theta}_{\pi}: \frac{c_{\pi}}{v_{\pi}},\tilde{D}_{\pi}:\text{otherwise} \}$  & Minimax Regret   \\ 
{SHED} (ours)  & $\Lambda(\pi)=\underset{\vec{\theta} \in \Theta}{\arg \max}Q_{\pi}(s=\pi,a=\vec{\theta})$ & Teacher Reward Maximization    \\ 
 \bottomrule
\end{tabular}
\caption{An overview of UED methods in relation to SHED. $U(\Theta)$ is a uniform distribution over environment parameter space, $\tilde{D}_{\pi}$ is a baseline distribution$, \bar{\theta}_{\pi}$ is the trajectory which maximizes regret of $\pi$, and $v_{\pi}$ is the value above the baseline distribution that $\pi$ achieves on that trajectory, $c_{\pi}$ is the negative of the worst-case regret of $\pi$. Details are described in PAIRED~\cite{dennis2020emergent}. 
}
\label{tab:teacher_choices}
\end{table*}

\section{Details about the Generative Model}\label{app:sec:generative_model}
\subsection{Generative model to generate synthetic next state}
Here, we describe how to leverage the diffusion model to learn the conditional data distribution in the collected experiences $\tau = \{(s^{u}_t, a^{u}_t, r^{u}_t, s^{u,\prime}_t) \}$. Later, we can use the trainable reverse chain in the diffusion model to generate the synthetic trajectories that can be used to help train the teacher agent, resulting in reducing the resource-intensive and time-consuming collection of upper-level teacher experiences. We deal with two different types of timesteps in this section: one for the diffusion process and the other for the upper-level teacher agent, respectively. We use subscripts $k \in {1,\dots, K}$ to represent diffusion timesteps and subscripts $t \in {1,\dots,T}$ to represent trajectory timesteps in the teacher's experience.

In the image domain, the diffusion process is implemented across all pixel values of the image. In our setting, we diffuse over the next state $s^{u,\prime}$ conditioned the given state $s^u$ and action $a^u$. We construct our generative model according to the conditional diffusion process:
\begin{equation*}
    q(s_{k}^{u, \prime}|s_{k-1}^{u, \prime}), \quad p_{\phi}(s_{k-1}^{u, \prime}|s_k^{u, \prime},s^{u},a^{u})
\end{equation*}
As usual, $ q(s_{k}^{u, \prime}|s_{k-1}^{u, \prime})$ is the predefined forward noising process while $p_{\phi}(s_{k-1}^{u, \prime}|s_k^{u, \prime},s^{u},a^{u})$ is the trainable reverse denoising process. We begin by randomly sampling the collected experiences $\tau = \{(s^{u}_t, a^{u}_t, r^{u}_t, s^{u,\prime}_t) \}$ from the real experience buffer $ {\mathcal{B}}_{real}$. 

We drop the superscript $u$ here for ease of explanation. Giving the observed state $s$ and action $a$, we use the reverse process $p_{\phi}$ to represent the generation of the next state $s^{\prime}$:
\begin{equation}\label{app:eq:reverse_chain_s0}
        p_{{\phi}}(s_{0:K}^{\prime}|s,a)=\mathcal{N}(s_K^{\prime};{0},\mathbf{I} ) \prod_{k=1}^K p_{\phi}(s^\prime_{k-1}|s_k^\prime,s, a)
\end{equation}
At the end of the reverse chain, the sample $s^{\prime}_0$, is the generated next state $s^\prime$.
As shown in Section~\ref{subsec:diff}, $p_{{\phi}}(s_{k-1}^{\prime}|,s_k^\prime,s, a)$ could be modeled as a Gaussian distribution $\mathcal{N} ( s_{k-1}^\prime ; \mu_{{\theta}}(s_k^\prime,s,a,k),\Sigma_{{\theta}}(s_k^\prime,s,a,k))$. Similar to \citet{ho2020denoising}, we parameterize $p_{\phi}(s^\prime_{k-1}|s_k^\prime,s, a)$ as a noise prediction model with the covariance matrix fixed as 
$$\Sigma_{{\theta}}(s_k^\prime,s,a,k)=\beta_i \mathbf{I}$$
and mean is 
$$\mu_{{\theta}}(s_i^\prime,s, a,k) =\frac{1}{\sqrt{\alpha_k}}\left( s_k^\prime - \frac{\beta_k}{\sqrt{1-\bar{\alpha}_k} }{\epsilon_{\theta}(s_k^\prime,s,a,k)} \right) $$

Where $\epsilon_{\theta}(s_k^\prime,s,a,k)$ is the trainable denoising function, which aims to estimate the noise $\epsilon$ in the noisy input $s_k^\prime$ at step $k$. Specifically, giving the sampled experience $(s,a,s^\prime)$, we begin by sampling $s_K^\prime \sim \mathcal{N}({0},\mathbf{I}) $ and then proceed with the reverse diffusion chain $p_{{\phi}}(s_{k-1}^{\prime}|,s_k^\prime,s, a)$ for $k=K,\dots, 1$. The detailed expression for $s_{k-1}^\prime$ is as follows:
\begin{equation}
    \frac{s_k^\prime}{\sqrt{\alpha_k}} - \frac{\beta_k}{\sqrt{\alpha_k(1-\bar{\alpha}_k})}{\epsilon_{{\theta}}(s_k^\prime,s,a,k)} + \sqrt{\beta_k}{\epsilon},
\end{equation}\label{eq:s_prime}
where ${\epsilon}\sim \mathcal{N}({0},\mathbf{I})$. Note that ${\epsilon}=0$ when $k = 1$.  

\paragraph{Training objective.} We employ a similar simplified objective, as proposed by~\citet{ho2020denoising} to train the conditional ${\epsilon}$- model through the following process:
\begin{equation}
    \mathcal{L}({\theta}) = 
    \mathbb{E}_{(s,a,s^\prime )\sim \tau,k \sim \mathcal{U},{\epsilon} \sim \mathcal{N}({0},\bf{I}) } \left[ \lVert {\epsilon}  - {\epsilon}_{{\phi}}( s^\prime_k,{s,a},k)  \rVert^2  \right] 
\end{equation}

Where $s^\prime_k=\sqrt{\bar{\alpha}_k}s^\prime + \sqrt{1-\bar{\alpha}_k }{\epsilon}$. $\mathcal{U}$ represents a uniform distribution over the discrete set $\{1,\dots, K\}$. The intuition for the loss function $\mathcal{L}({\theta})$ tries to predict the noise $\epsilon\sim \mathcal{N}(0,\mathbf{I})$ at the denoising step $k$, and the diffusion model is essentially learning the student policy involution trajectories collected in the real experience buffer $\mathcal{B}_{reals}$. Note that the reverse process necessitates a substantial number of steps $K$, as the Gaussian assumption holds true primarily under the condition of the infinitesimally limit of small denoising steps~\citep{sohl2015deep}. Recent research by~\citet{xiao2021tackling} has demonstrated that enabling denoising with large steps can reduce the total number of denoising steps $K$. To expedite the relatively slow reverse sampling process outlined in Equation~\ref{app:eq:reverse_chain_s0} (as it requires computing ${\epsilon}_{{\phi}}$ networks $K$ times), we use a small value of $K$, while simultaneously setting $\beta_{\min} = 0.1$ and $\beta_{\max} = 10.0$. Similar to~\citet{wang2022diffusion}, we define:
\begin{equation*}
\begin{aligned}
    \beta_k &=1-\alpha_k \\
    &=1-\exp\left(\beta_{\min}\times \frac{1}{K} - 0.5(\beta_{\max}-\beta_{\min})\frac{2k -1}{K^2}\right)
\end{aligned}
\end{equation*}
This noise schedule is derived from the variance-preserving Stochastic Differential Equation by~\citet{song2020score}. 

\paragraph{Generate Synthetic Trajectories.} Once the diffusion model has been trained, it can be used to generate synthetic experience data by starting with a draw from the prior $s_K^\prime\sim \mathcal{N}({0},\mathbf{I})$ and successively generating denoised next state, conditioned on the given $s$ and $a$ through the reverse chain $p_{\phi}$ in Equation~\ref{app:eq:reverse_chain_s0}. Note that the given condition action $a$ can either be randomly sampled from the action space (which is also the environment parameter space) or use another diffusion model to learn the action distribution giving the initial state $s$. In such case, this new diffusion model is essentially a behavior-cloning model that aims to learn the teacher policy $\Lambda(a|s)$. This process is similar to the work of~\citet{wang2022diffusion}. We discuss this process in detail in the appendix. In our implementation, we opt for random action sampling to maintain simplicity and increase synthetic experience diversity, ultimately contributing to a more robust teacher agent.
  
\begin{figure*}[t]
    \centering
    \begin{minipage}[b]{0.32\linewidth}
        \includegraphics[width=\linewidth]{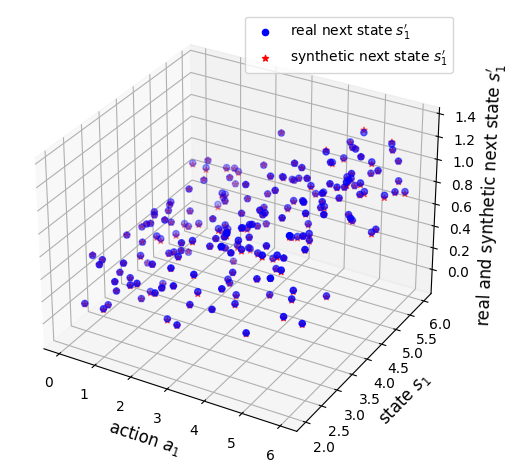}
    \end{minipage}
    \hfill
    \begin{minipage}[b]{0.32\linewidth}
        \includegraphics[width=\linewidth]{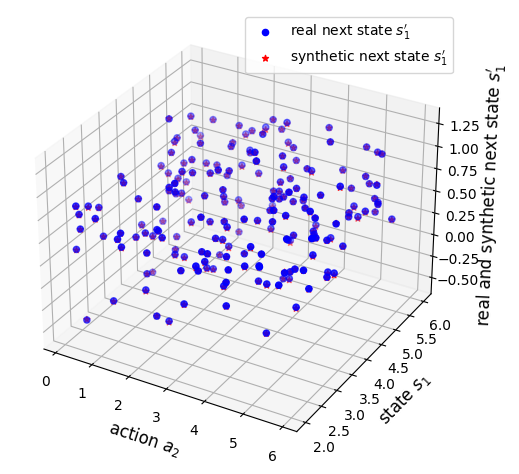}
    \end{minipage}
    \hfill
    \begin{minipage}[b]{0.32\linewidth}
    \includegraphics[width=\linewidth]{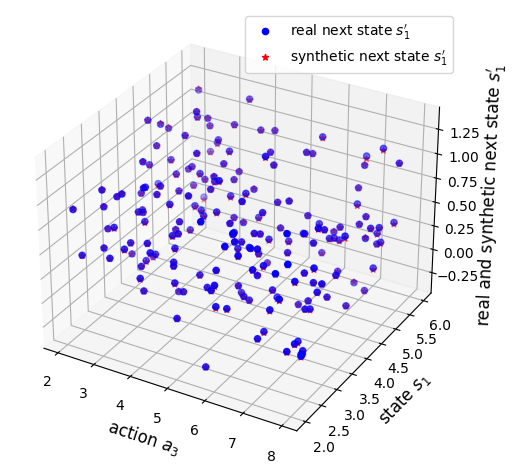}
    \end{minipage}
    \hfill
    \begin{minipage}[b]{0.32\linewidth}
    \includegraphics[width=\linewidth]{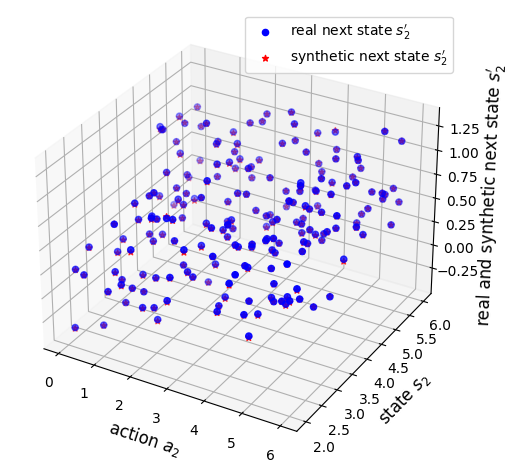}
    \end{minipage}
    \hfill
    \begin{minipage}[b]{0.32\linewidth}
    \includegraphics[width=\linewidth]{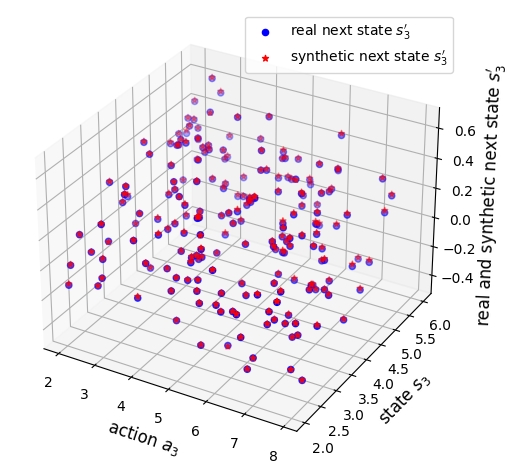}
    \end{minipage}
    \hfill
    \begin{minipage}[b]{0.32\linewidth}
    \includegraphics[width=\linewidth]{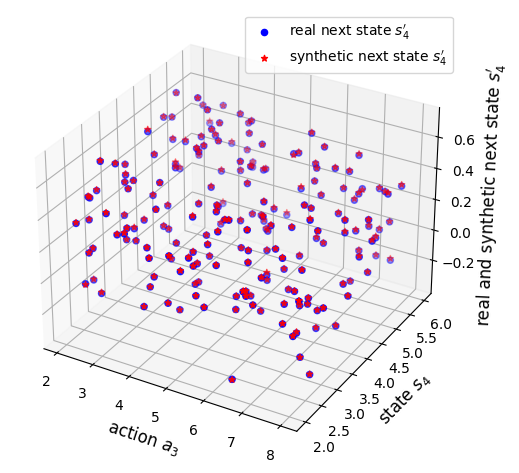}
    \end{minipage}
    \caption{The distribution of the real $s^\prime$ and the synthetic $s^\prime$ conditioned on $(s,a)$.}
    \label{app:fig:diffusion_small_noise}
\end{figure*}
\subsection{Generative model to generate synthetic action}
Once the diffusion model has been trained, it can be used to generate synthetic experience data by starting with a draw from the prior $s_K^\prime\sim \mathcal{N}({0},\mathbf{I})$ and successively generating denoised next state, conditioned on the given $s$ and $a$ through the reverse chain $p_{\phi}$ in Equation~\ref{app:eq:reverse_chain_s0}. Note that the giving condition action $a$ can either be randomly sampled from the action space (which is also the environment parameter space) or we can train another diffusion model to learn the action distribution giving the initial state $s$, and then use the trained new diffusion model to sample the action $a$ giving the state $s$. This process is similar to the work of~\citet{wang2022diffusion}. 

In particular, we construct another conditional diffusion model as:
\begin{equation*}
    q(a_{k}|a_{k-1}), \quad p_{\phi}(a_{k-1}|a_k,s)
\end{equation*}
As usual, $q(a_{k}|a_{k-1})$ is the predefined forward noising process while $p_{\phi}(a_{k-1}|a_k,s)$ is the trainable reverse denoising process. we represent the action generation process via the reverse chain of the conditional diffusion model as 
\begin{equation}\label{app:eq:reverse_chain_a0}
        p_{{\phi}}(a_{0:K}|s)=\mathcal{N}(a_K;{0},\mathbf{I} ) \prod_{k=1}^K p_{\phi}(a_{k-1}|a_k,s)
\end{equation}
At the end of the reverse chain, the sample $a_0$, is the generated action $a$ for the giving state $s$.
Similarly, we parameterize $p_{\phi}(a_{k-1}|a_k,s)$ as a noise prediction model with the covariance matrix fixed as 
$$\Sigma_{{\theta}}(a_k,s,k)=\beta_i \mathbf{I}$$
and mean is 
$$\mu_{{\theta}}(a_i,s, k) =\frac{1}{\sqrt{\alpha_k}}\left( a_k - \frac{\beta_k}{\sqrt{1-\bar{\alpha}_k} }{\epsilon_{\theta}(a_k,s,k)} \right) $$

Similarly, the simplified loss function is 
\begin{equation}
    \mathcal{L}^{a}({\theta}) = 
    \mathbb{E}_{(s,a )\sim \tau,k \sim \mathcal{U},{\epsilon} \sim \mathcal{N}({0},\bf{I}) } \left[ \lVert {\epsilon}  - {\epsilon}_{{\phi}}( a_k,{s},k)  \rVert^2  \right] 
\end{equation}

Where $a_k=\sqrt{\bar{\alpha}_k}a + \sqrt{1-\bar{\alpha}_k }{\epsilon}$. $\mathcal{U}$ represents a uniform distribution over the discrete set $\{1,\dots, K\}$. The intuition for the loss function $\mathcal{L}^{a}({\theta})$ tries to predict the noise $\epsilon\sim \mathcal{N}(0,\mathbf{I})$ at the denoising step $k$, and the diffusion model is essentially a behavior cloning model to learn the student policy collected in the real experience buffer $\mathcal{B}_{reals}$.

Once this new diffusion model is trained, the generation of the synthetic experience can be formulated as: 
\begin{itemize}
    \item we first randomly sample the state from the collected real trajectories $s\sim \tau$;
    \item we use the new diffusion model discussed above to mimic the teacher's policy to generate the actions $a$;
    \item giving the state $s$ and action $a$, we use the first diffusion model presented in the main paper to generate the next state $s^\prime$;
    \item we compute the reward $r$ according to the reward function, and add the final generated synthetic experience $(s,a,r,s^\prime)$ to the synthetic experience buffer $\mathcal{B}_{syn}$ to help train the teacher agent.
\end{itemize}

\section{Empirical Analysis of Generative Model}\label{app:sec:diffusion}

\subsection{Ability to Generate Good Synthetic Trajectories}
We investigate SHED's ability to assist in collecting experiences for the upper-level MDP teacher. This involves the necessity for SHED to prove its ability to accurately generate synthetic experiences for teacher agents. To check the quality of these generated synthetic experiences, we employ a diffusion model to simulate some data for validation (even though Diffusion models have demonstrated remarkable success across vision and NLP tasks).

We design the following experiment: given the teacher's observed state $s^u=[p_1,p_2,p_3,p_4,p_5]$, where $p_i$ denotes the student performance on $i$-th evaluation environment. and given the teacher's action $a^u=[a_1,a_2,a_3]$, which is the environment parameters and are used to generate corresponding environment instances. We use a neural network $f(s^u,a^u)$ to mimic the involution trajectories of the student policy $\pi$. That is, with the input of the state $s^u$ and action $a^u$ into the neural network, it outputs the next observed state $s^{u,\prime} = [p_1^\prime,p_2^\prime,p_3^\prime,p_4^\prime,p_5^\prime]$, indicating the updated student performance vector on the evaluation environments after training in the environment generated by $a^u$. In particular, we add a noise $\varepsilon$ into $s^{u,\prime}$ to represent the uncertainty in the transition. We first train our diffusion model on the real dataset $(s^u, a^u, s^{u,\prime})$ generated by neural network $f(s^u, a^u)$. We then set a fixed $(s^u, a^u)$ pair and input them into $f(s^u, a^u)$ to generate 200 samples of real $s^{u,\prime}$. The trained diffusion model is then used to generate 200 synthetic $s^{u,\prime}$ conditioned on the fixed $(s^u, a^u)$ pair.

\begin{figure*}[t]
    \centering
    \begin{minipage}[b]{0.32\linewidth}
        \includegraphics[width=\linewidth]{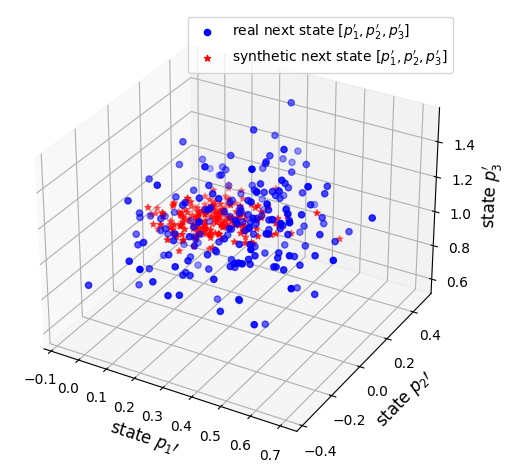}
    \end{minipage}
    \hfill
    \begin{minipage}[b]{0.32\linewidth}
        \includegraphics[width=\linewidth]{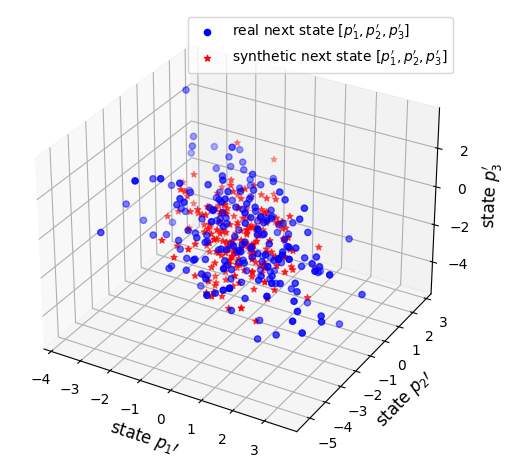}
    \end{minipage}
    \hfill
    \begin{minipage}[b]{0.32\linewidth}
    \includegraphics[width=\linewidth]{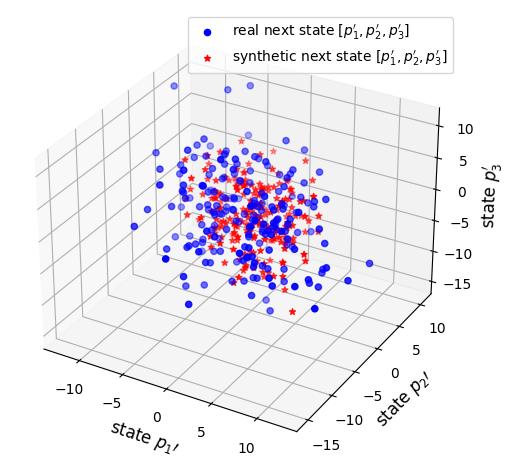}
    \end{minipage}
    \caption{The distribution of the real $[s^\prime_1,s^\prime_2,s^\prime_3]$(red) and the synthetic $[s^\prime_1,s^\prime_2,s^\prime_3]$(blue) giving the fixed $(s^u,a^u)$. Specifically, the noise $\varepsilon$ in $f(s^u, a^u)$ is (i).{\em left} figure: $\varepsilon=\epsilon$, (ii).{\em middle} figure: $\varepsilon=3*\epsilon$, (iii).{\em right} figure: $\varepsilon=10*\epsilon$, where $\epsilon~\sim \mathcal{N}(0,1)$.} 
    \label{fig:diffusion}
\end{figure*}

The results are presented in Figure~\ref{fig:diffusion}, we can see that the generative model can effectively capture the distribution of real experience even if there is a large uncertainty in the transition, indicated by the value of $\varepsilon$. This provides evidence that the diffusion model can generate useful experiences conditioned on $(s^u,a^u)$. It is important to note that the marginal distribution derived from the reverse diffusion chain provides an implicit, expressive distribution, such distribution has the capability to capture complex distribution properties, including skewness and multi-modality.

\subsection{Addition Experiments on Diffusion Model}\label{app:ssec:different_setting}
We further provide more results to show the ability of our generative model to generate synthetic trajectories where the noise is extremely small. In such cases, the actual next state $s^\prime$ will converge to a certain value, and the synthetic next state $s^{syn, \prime}$ generated by the diffusion model should also be very close to that value, then the diffusion model has the ability to sample the next state $s_0^{syn, \prime}$ which can accurately represent the next state. We present the results in Figure~\ref{app:fig:diffusion_small_noise}. Specifically, this figure shows when the noise is very small in the actual next state, which is $0.05*\epsilon$, and $\epsilon\sim \mathcal{N}(0,1)$. Giving any condition $(s,a)$ pair, we selectively report on $(s_i, a_i)$, where $x$-axis is the $a_i$ value, and $y$-axis is the $s_i$ value. The student policy with initial performance vector $s$ is trained on the environments generated by the teacher's action $a$. We report the new performance $s_i^\prime$ of student policy on $i$-th environments after training in the $z$-axis. In particular, if two points $s^\prime_i$ and $s^{syn, \prime}_i$ are close, it indicates that the diffusion model can successfully generate the actual next state. As we can see, when the noise is extremely small, our diffusion model can accurately predict the next state of $s_i^\prime$ giving any condition $(s,a)$ pair. 

\begin{figure*}[t]
    \centering
    \begin{minipage}[b]{0.46\linewidth}
        \includegraphics[width=\linewidth]{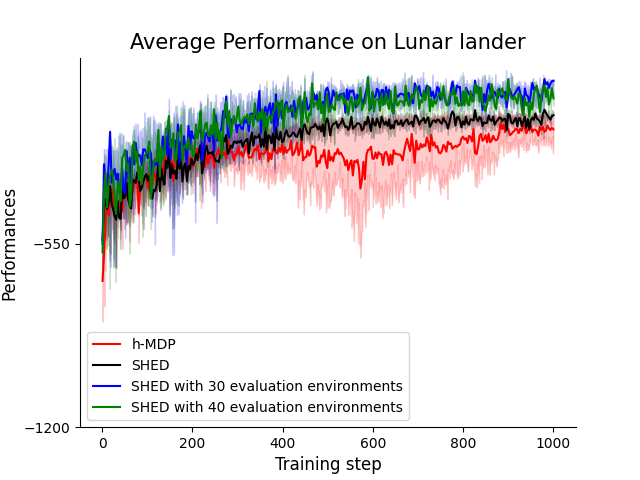}
    \end{minipage}
    \hfill
    \begin{minipage}[b]{0.46\linewidth}
        \includegraphics[width=\linewidth]{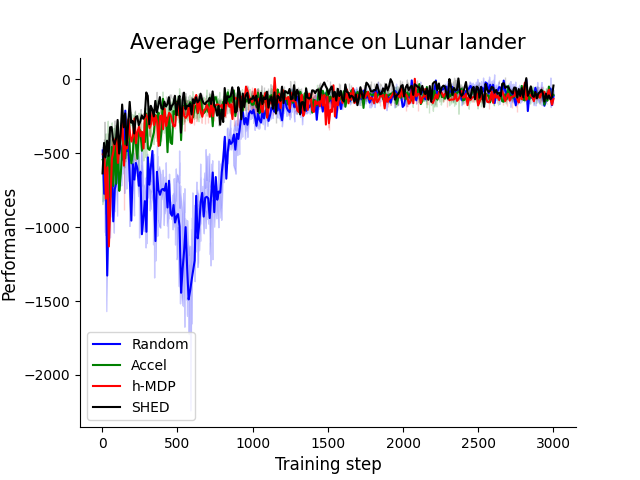}
    \end{minipage}
    \caption{{\em Left}: The ablation study in the Lunar lander environment which investigates the effect of the size of the evaluation environment set from 10 to 30 and 40. We provide the average zero-shot transfer performances on the test environments (mean and standard error). {\em Right}: Zero-shot transfer performance on the test environments under a longer time horizon in Lunar lander environments(mean and standard error).}
    \label{app:fig:ablation_performance}
\end{figure*}

\section{Additional Experiment Details}
\subsection{ Hyperparameters}\label{app:Hyperparameters}
We set the learning rate $1e-3$ for actor, and $3e-3$ for critic, we set gamma $\gamma=0.999$, $\lambda=0.95$, and set coefficient for the entropy bonus (to encourage exploration) as $0.01$. For each environment, we conduct 50 PPO updates for the student agent, and We can train on up to 50 environments, including replay. For our diffusion model, the diffusion discount is 0.99, and batch size is 64, $\tau$ is 0.005, learning rate is $3e-4$. The synthetic buffer size is $1000$, and the ratio is 0.25.

\subsection{Experiments Compute Resources}\label{app:gpu}
All the models were trained on a single NVIDIA GeForce RTX 3090 GPU and 16 CPUs.

\subsection{Maze Document}\label{app:ssub:maze_doc}

Here we provide the document with the instruction to generate feasible maze environments.

\begin{lstlisting}
There are several factors that can affect the difficulty of a maze. Here are some key factors to consider:

1. Maze Size: Larger mazes generally increase complexity and difficulty as the agent has more states to explore. Typically, maze size should be $>4*4$ and $<15*15$.
- Size $\leq 7*7$: easy
- Size >7*7 but <10*10: medium
- Size >10x10 but <15*15: hard

2. Maze Structure: Path complexity, including twists, turns, and dead-ends, impacts navigation strategies. Narrow corridors vs wide spaces also matter.
- $<2$ turns in feasible path from start to end: easy
- $2-4$ turns in path from start to end: medium  
- $\geq 4$ turns in path from start to end: hard

3. Goal Location: Distance from start to end affects difficulty.
- Path requires <5 steps: easy
- Path requires 5-10 steps: medium
- Path requires >10 steps: hard

4. Start Location: Starting position categories:
- Level 1: near top left
- Level 2: near top right  
- Level 3: near bottom left
- Level 4: near bottom right
- Level 5: near center

Note: Maze uses: -1=blocks, 0=feasible path, 1=start, 2=end.
Must have feasible path (1 and 2 connected via 0s or directly)!

Example Feasible Maze:
[
  [0, -1, -1, 2],
  [1, -1, 0, 0],
  [0, -1, 0, -1], 
  [0, 0, 0, -1],
]

Non-Feasible Examples:
[
  [0, -1, -1, 2],
  [1, -1, 0, 0],
  [0, -1, -1, 0],
  [0, 0, 0, -1],
]
Or
[
  [1, -1],
  [-1, 2]
]
\end{lstlisting}

\subsection{Prompt for RAG}\label{app:ssub:prompt}
We provide our prompt for the Retrieval Augmented Generation as follows:

\begin{lstlisting}
Please refer to the document, and generate a maze with feasible path. 
The difficulty level for the maze size is {maze_size_level}, and the 
difficulty level for the maze structure is {maze_structure_level}, 
the difficulty level for the goal location is {goal_location_level}, 
the difficulty level for the start location is {start_position_level}.
\end{lstlisting}

\section{Additional experiments}\label{app:ablation_addition_exp}

\subsection{Ablation Study}\label{app:ssec:ablation}
We also provide ablation analysis to evaluate the impact of different design choices in Lunar lander domain, including (a) a larger evaluation environment set; (b) a bigger budget for constraint on the number of generated environments (which incurs a longer training time horizon). 
The results are reported in Figure~\ref{app:fig:ablation_performance}.

We examined the impact of the diffusion model and evaluation environment set size on performance. As shown in Figure~\ref{app:fig:ablation_performance} (left), SHED consistently outperforms h-MDP across all configurations, demonstrating the effectiveness of our generative model in enhancing teacher policy training through synthetic experiences.

We also observed that increasing the evaluation environment set size generally improves zero-shot transfer performance. This aligns with Theorem~\ref{thm:cover}, as a larger, more diverse evaluation set provides a better approximation of the student policy. However, we noted that SHED with 30 evaluation environments slightly outperformed the configuration with 40 environments. This unexpected result likely stems from the increased dimensionality of the student performance vector with 40 environments, which creates greater challenges for training an effective diffusion model given our limited dataset size.

We conduct experiments in Lunar lander under a longer time horizon. The results are provided on the right side of Figure~\ref{app:fig:ablation_performance}. As we can see, our proposed algorithm SHED can efficiently train the student agent to achieve the general capability in a shorter time horizon,  This observation indicates that our proposed environment generation process can better generate the suitable environments for the current student policy,  thereby enhancing its general capability, especially when there is a constraint on the number of generated environments.

\subsection{Impact of Evaluation Environment Selection}\label{app:subsec:div_eval_envs}
\begin{figure*}[th]
    \centering
    \includegraphics[width=\linewidth]{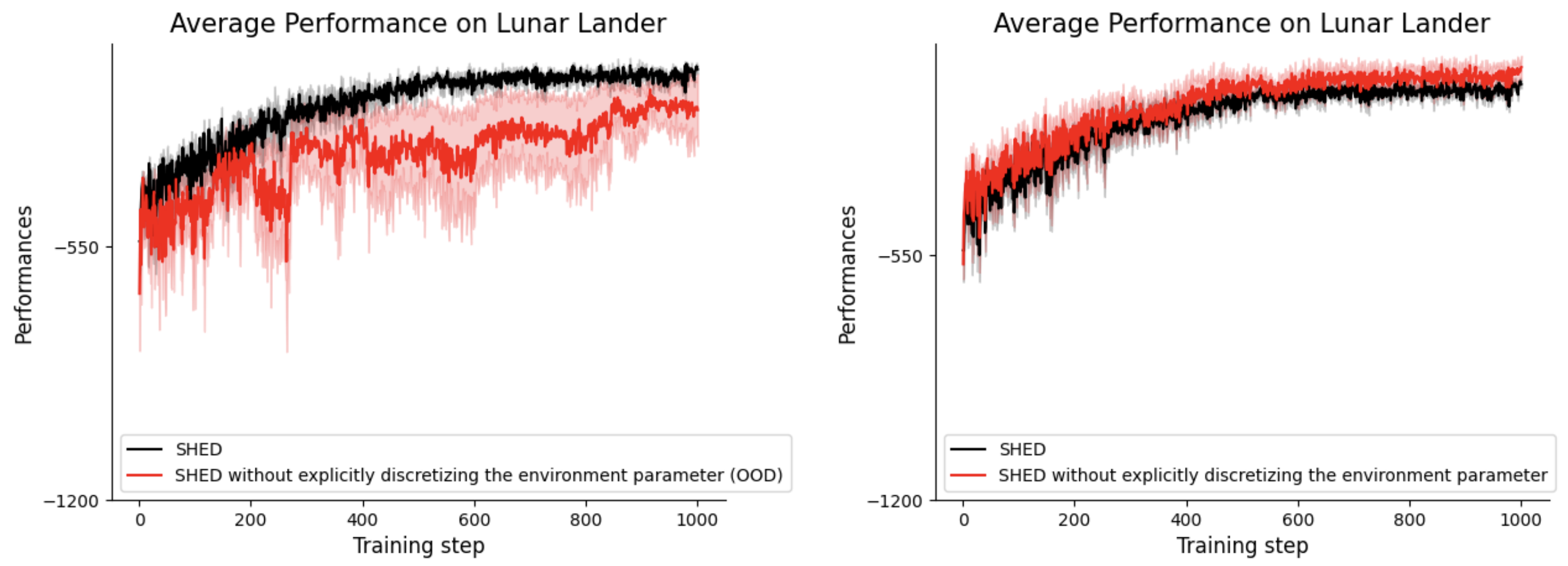}
    \caption{The average zero-shot transfer performances on the test environments in the Lunar lander environment (mean and standard error)Left: The evaluation environment does not explicitly discretize environment parameters and includes some Out-of-Distribution environments (very challenging environments). Right: The evaluation environment does not explicitly discretize environment parameters; instead, it includes only some in-distribution environments (solvable environments).}\label{fig:lander_ood}
\end{figure*}

We investigated how evaluation environment selection affects SHED's ability to generate effective curricula. Figure~\ref{fig:lander_ood} presents two conditions: environments with out-of-distribution (OOD) parameters that create infeasible scenarios (left), and environments with controlled parameters without discretization (right).

Our results show that using random levels without parameter discretization can enhance performance when the parameters remain reasonable and avoid OOD values (right panel). However, when evaluation environments predominantly include complex, unreasonable parameters (OOD), performance deteriorates significantly (left panel).

These findings highlight that diversity in evaluation environments is essential, but parameters must remain within reasonable distributions to ensure stable and effective performance. This underscores the importance of thoughtful evaluation environment selection when implementing SHED in new domains.

\subsection{Additional Lunar Lander Experiments}\label{app:ssec:lunar_lander_add}
\begin{figure}[th]
    \centering
    \includegraphics[width=0.8\linewidth]{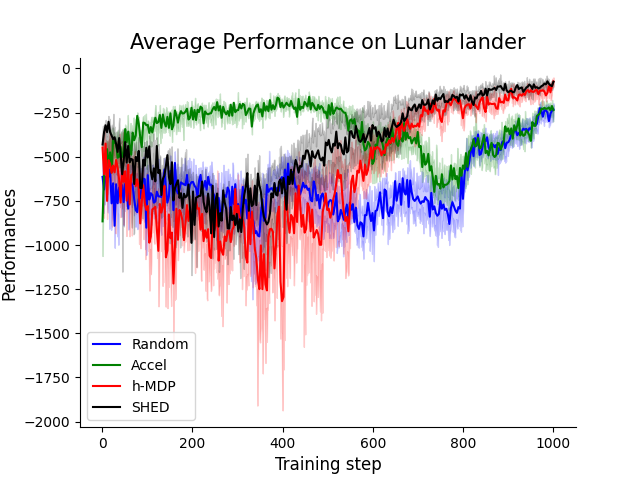}
    \caption{Zero-shot transfer performance on the test environments with a larger $cv$ value coefficient in Lunar lander environments. }
    \label{app:fig:lander_cv}
\end{figure}
\paragraph{Impact of Fairness Reward Weighting}

We conducted experiments to evaluate SHED's performance with a higher coefficient on the fairness reward component ($\eta=10$). As shown in Figure~\ref{app:fig:lander_cv}, this configuration leads to an interesting pattern: when fairness is heavily weighted, the teacher initially generates environments that cause a rapid performance decline followed by consistent improvement across all test environments.

This strategy appears to help the teacher maximize cumulative reward by avoiding substantial negative fairness penalties later in training. The teacher prioritizes balanced improvement across all environments rather than allowing some to improve at the expense of others. Importantly, despite this initial performance drop, students trained with high fairness weighting still achieved final performance surpassing all baseline methods.

\begin{figure*}[t]
\centering
\subfigure{\includegraphics[width=0.49\textwidth]{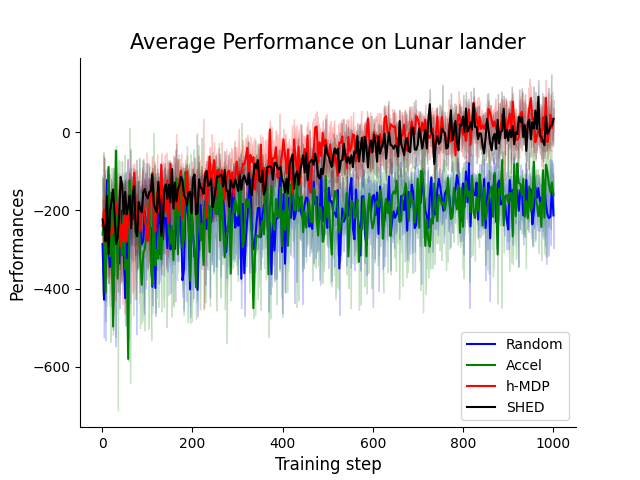}}
\subfigure{\includegraphics[width=0.49\textwidth]{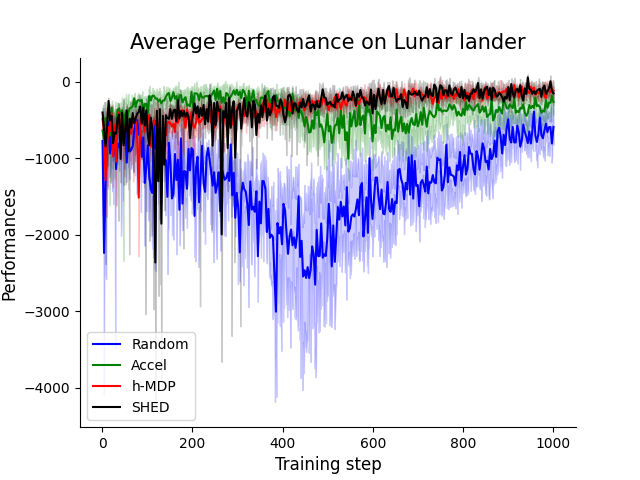}}
\subfigure{\includegraphics[width=0.49\textwidth]{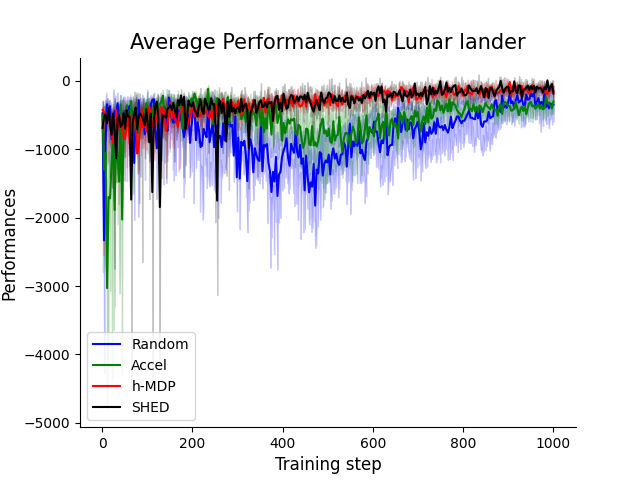}}
\subfigure{\includegraphics[width=0.49\textwidth]{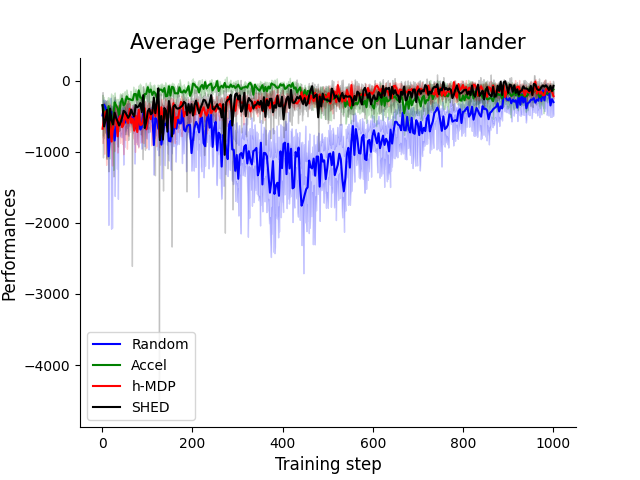}}
\subfigure{\includegraphics[width=0.49\textwidth]{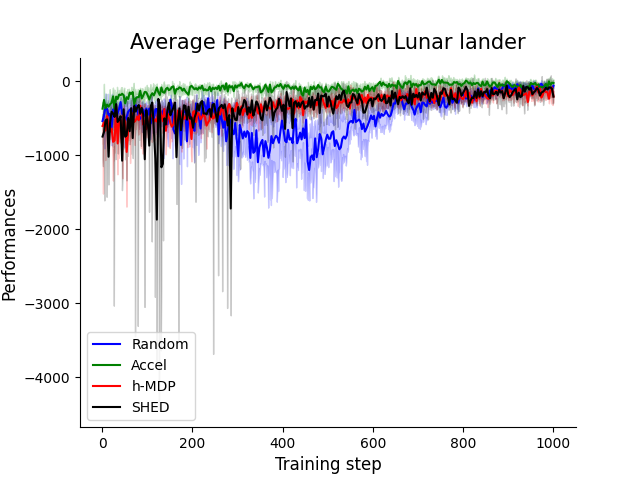}}
\subfigure{\includegraphics[width=0.49\textwidth]{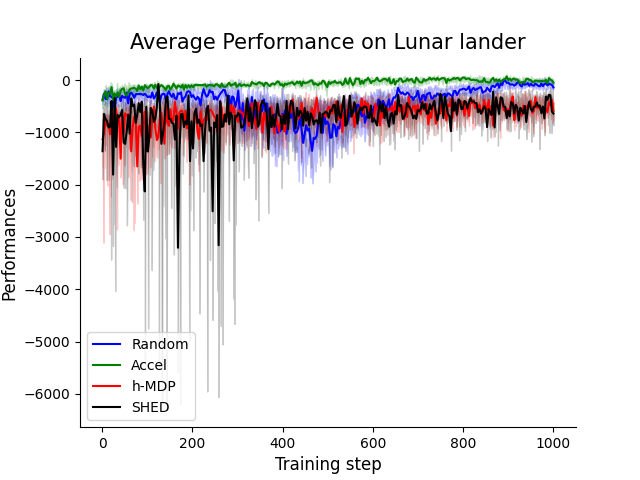}}
\caption{Performance curves for selected test environments for Lunar Lander. Note that ACCEL performs better in some environments, whereas SHED outperforms ACCEL in others. We reported the aggregate of all test environments in our main paper.}
\label{fig:10landers}
\end{figure*}

\begin{figure*}[t]
\centering
\subfigure{\includegraphics[width=0.49\textwidth]{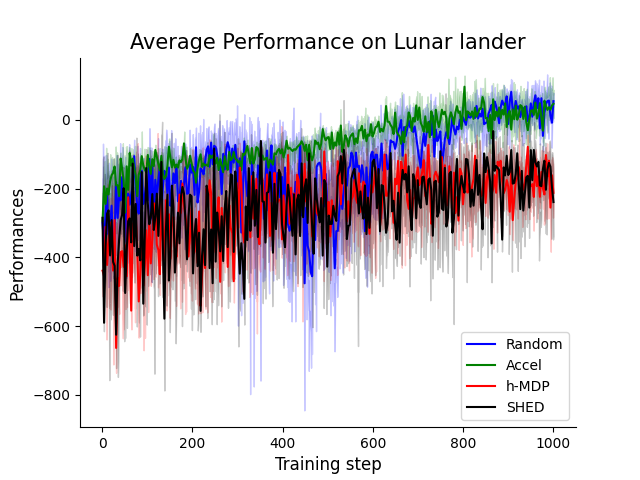}}
\subfigure{\includegraphics[width=0.49\textwidth]{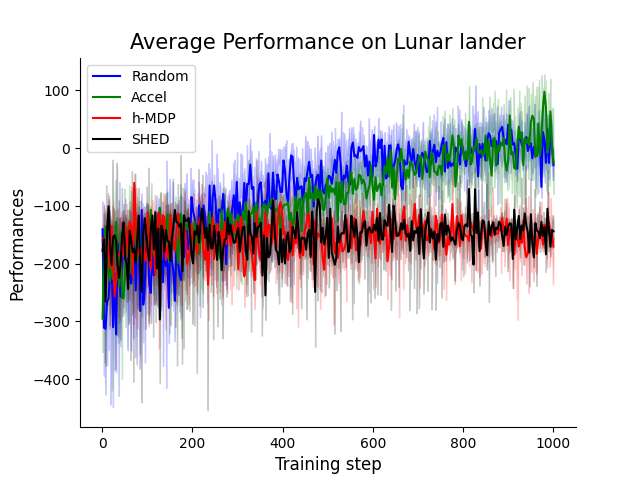}}
\subfigure{\includegraphics[width=0.49\textwidth]{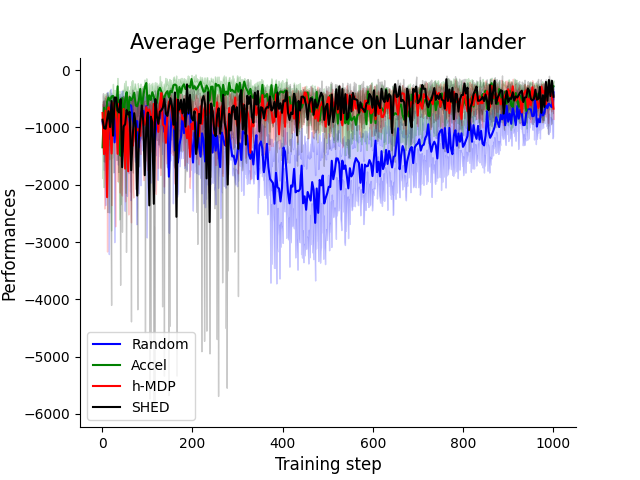}}
\subfigure{\includegraphics[width=0.49\textwidth]{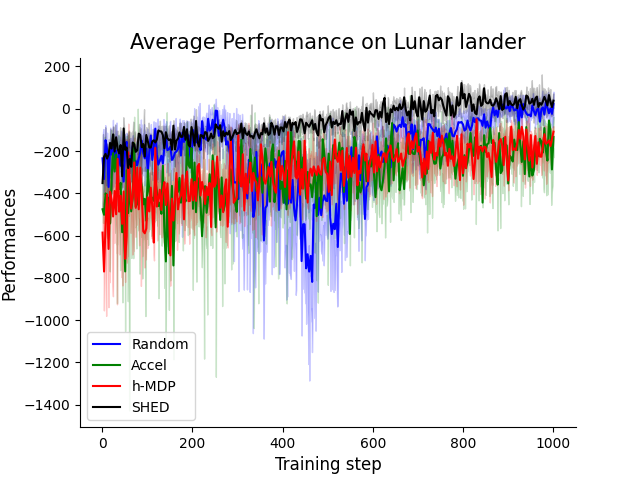}}
\caption{erformance curves for selected test environments for Lunar Lander, a continuation to Figure \ref{fig:10landers}}
\label{fig:10landers_2}
\end{figure*}

Detailed performance trajectories for each test environment are shown in Figures~\ref{fig:10landers} and~\ref{fig:10landers_2}. These graphs reveal that SHED and ACCEL compete closely for top performance in some individual test environments, highlighting the competitiveness of both algorithms. In our main results, we report aggregate performance across all test environments, which demonstrates SHED's overall superior performance despite this environment-specific variation.

\subsection{Additional Experiments on Maze}\label{app:ssec:maze}
We report some results of zero-shot transfer performances in maze test environments. The results are provided in Figure 12-16.

\begin{figure}[t]
 \centering
 \begin{minipage}[b]{0.48\textwidth}
   \centering
   \includegraphics[width=\linewidth]{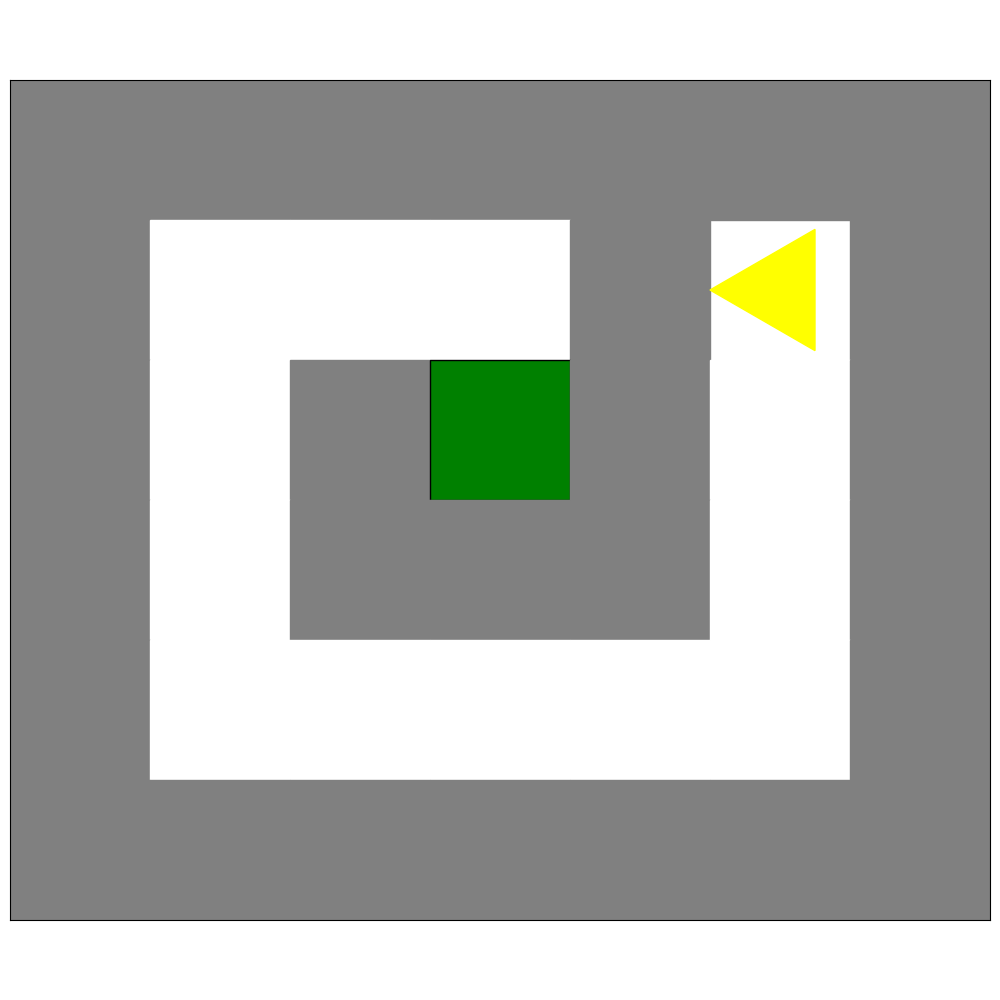}\\
   \includegraphics[width=\linewidth]{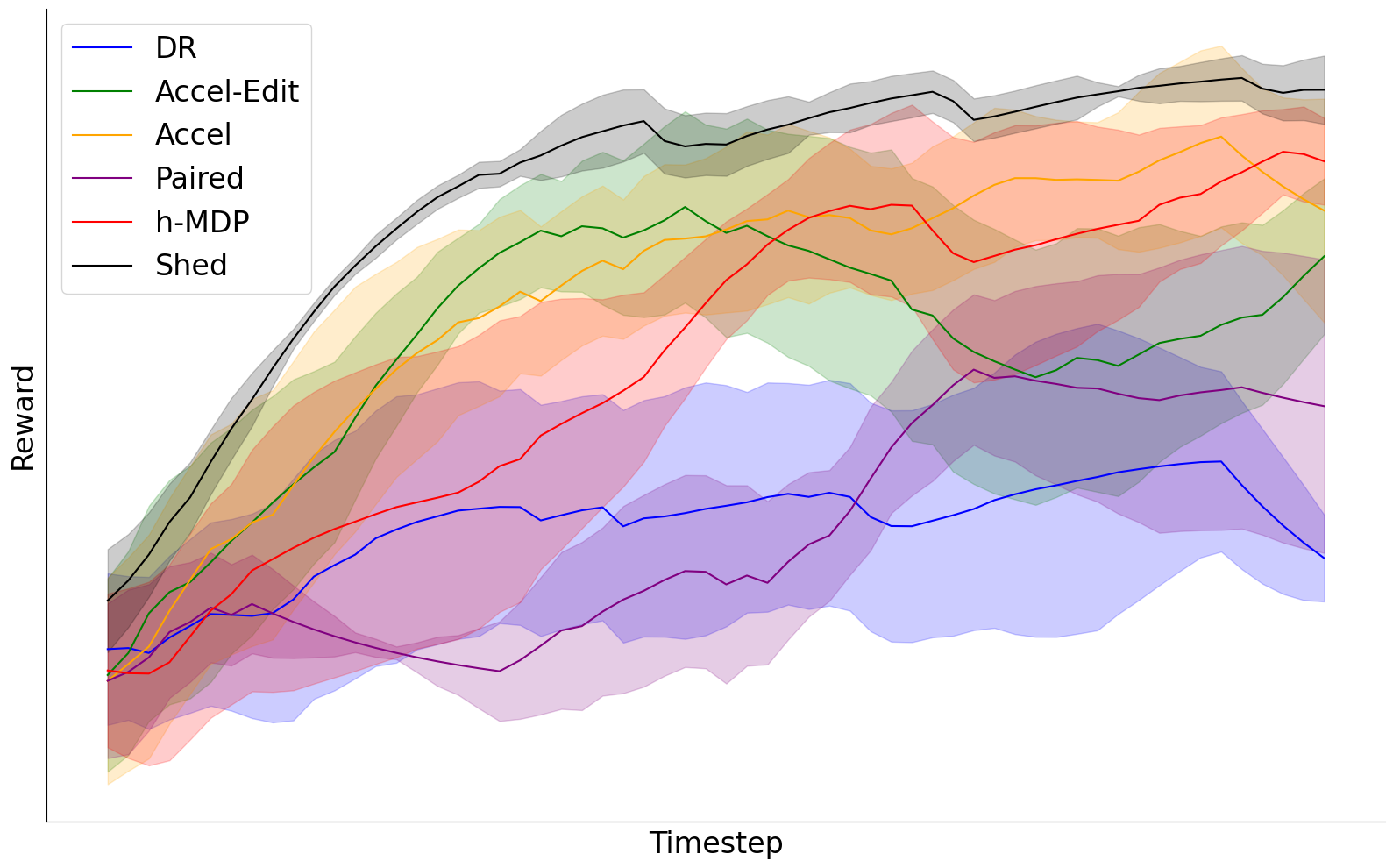}
   \caption{Small maze environment }
   \label{fig:small_maze}
 \end{minipage}
 \hfill
 \begin{minipage}[b]{0.48\textwidth}
   \centering
   \includegraphics[width=\linewidth]{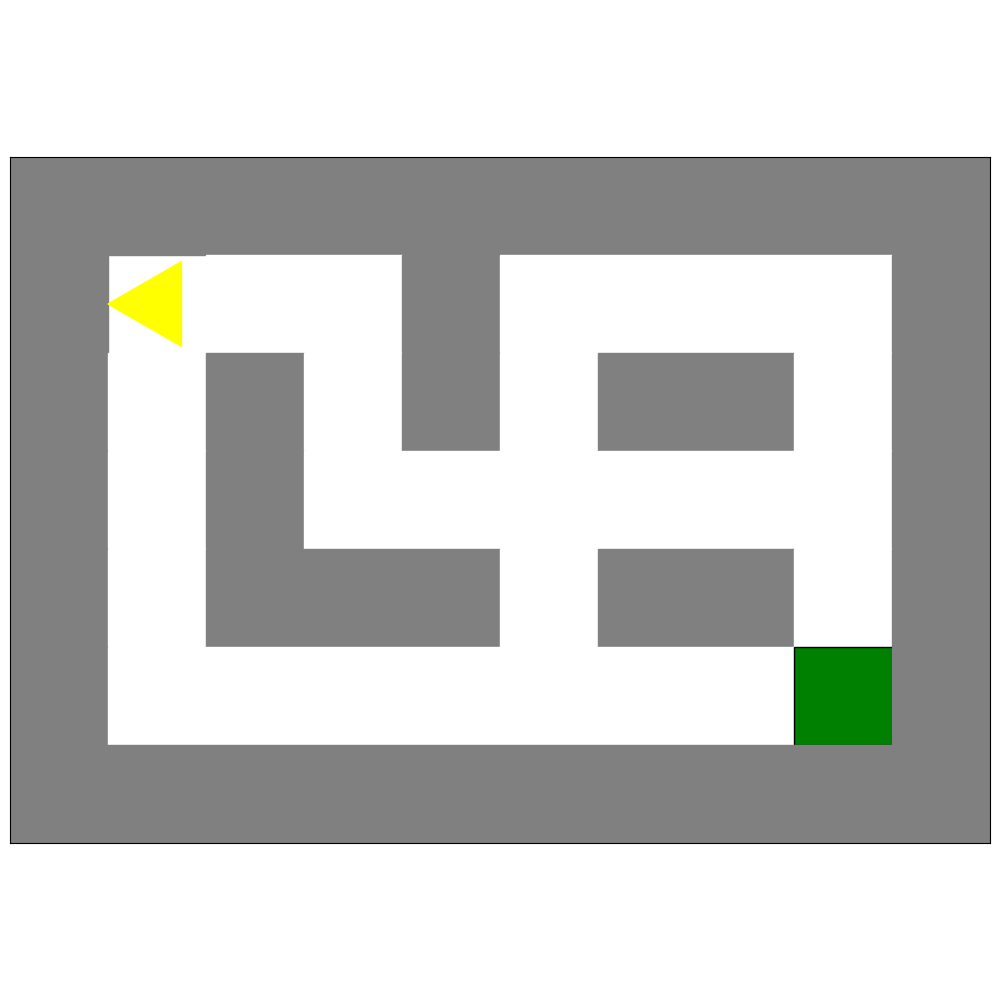}\\
   \includegraphics[width=\linewidth]{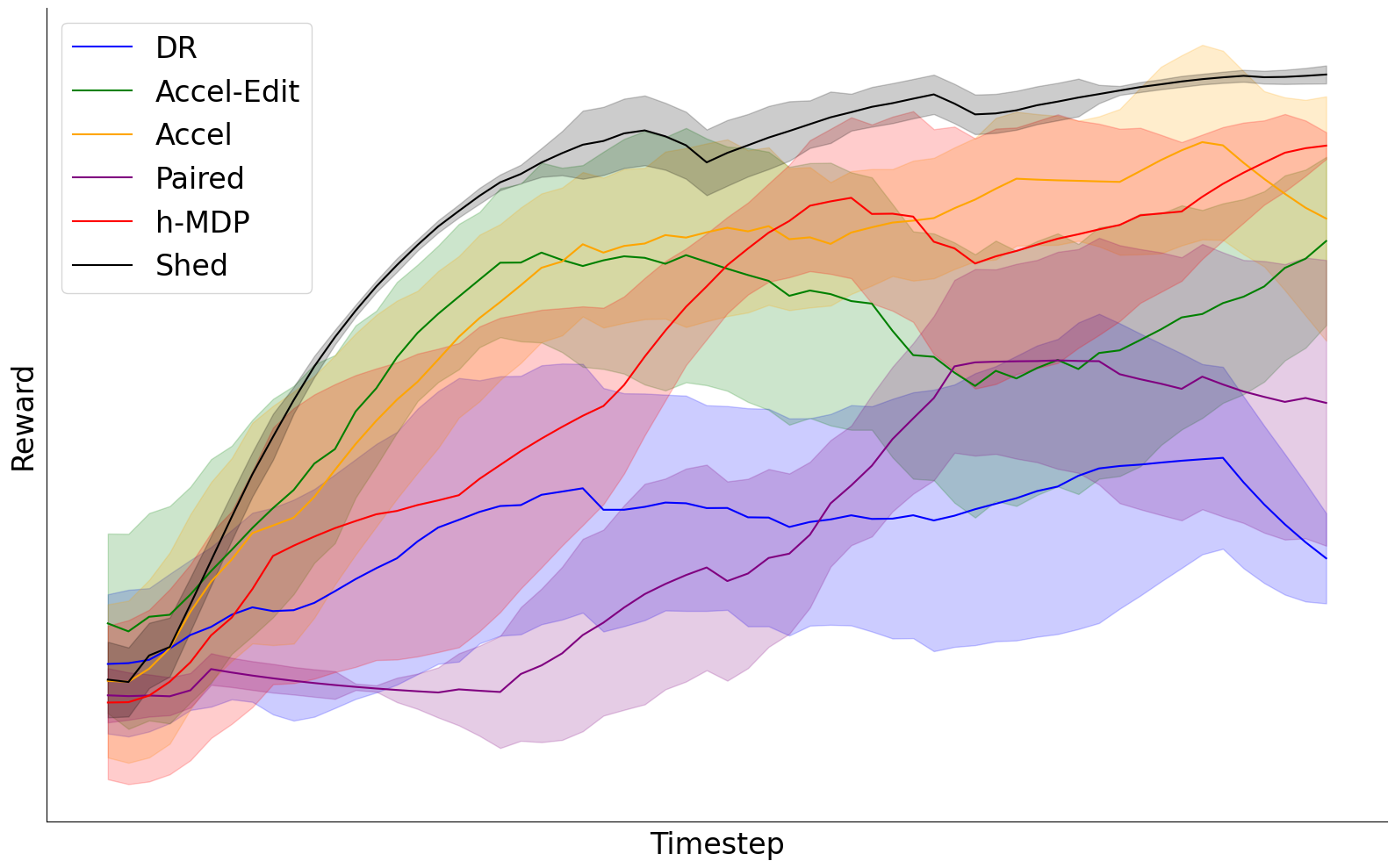}
   \caption{Medium maze environment }
   \label{fig:medium_maze}
 \end{minipage}
 
 \vspace{10pt}
 
 \begin{minipage}[b]{0.48\textwidth}
   \centering
   \includegraphics[width=\linewidth]{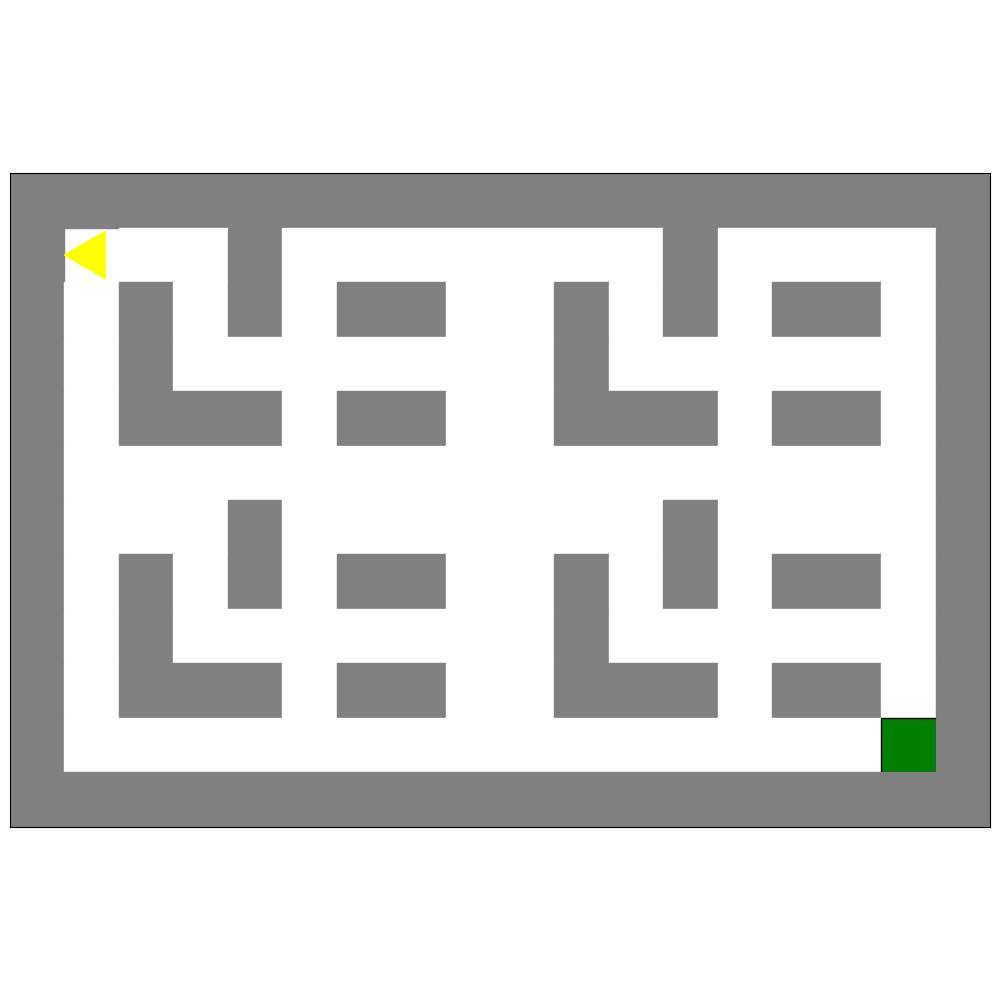}\\
   \includegraphics[width=\linewidth]{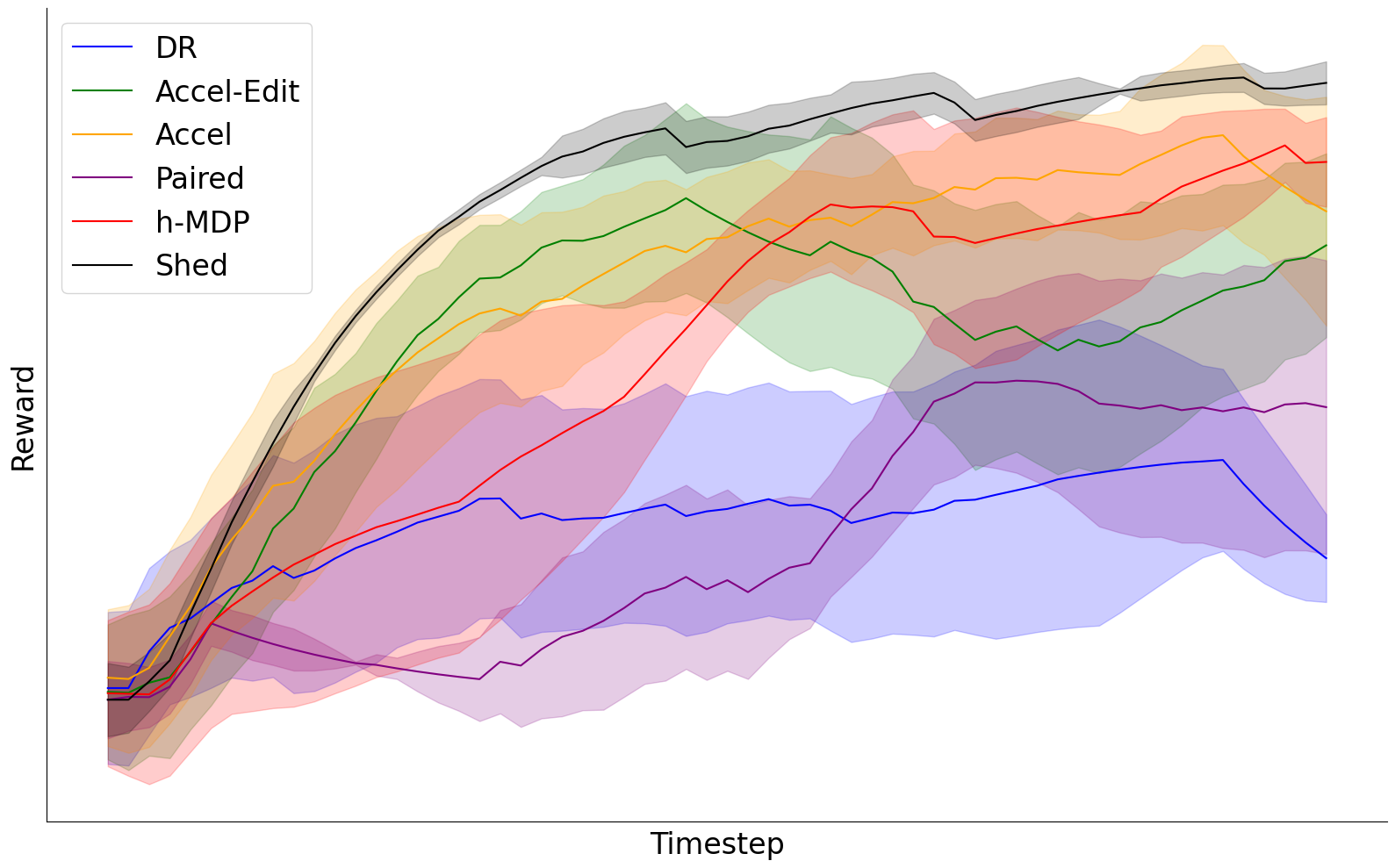}
   \caption{Large maze environment }
   \label{fig:large_maze}
 \end{minipage}
\end{figure}

\begin{figure}[t]
 \centering
 \begin{minipage}[b]{0.48\textwidth}
   \centering
   \includegraphics[width=\linewidth]{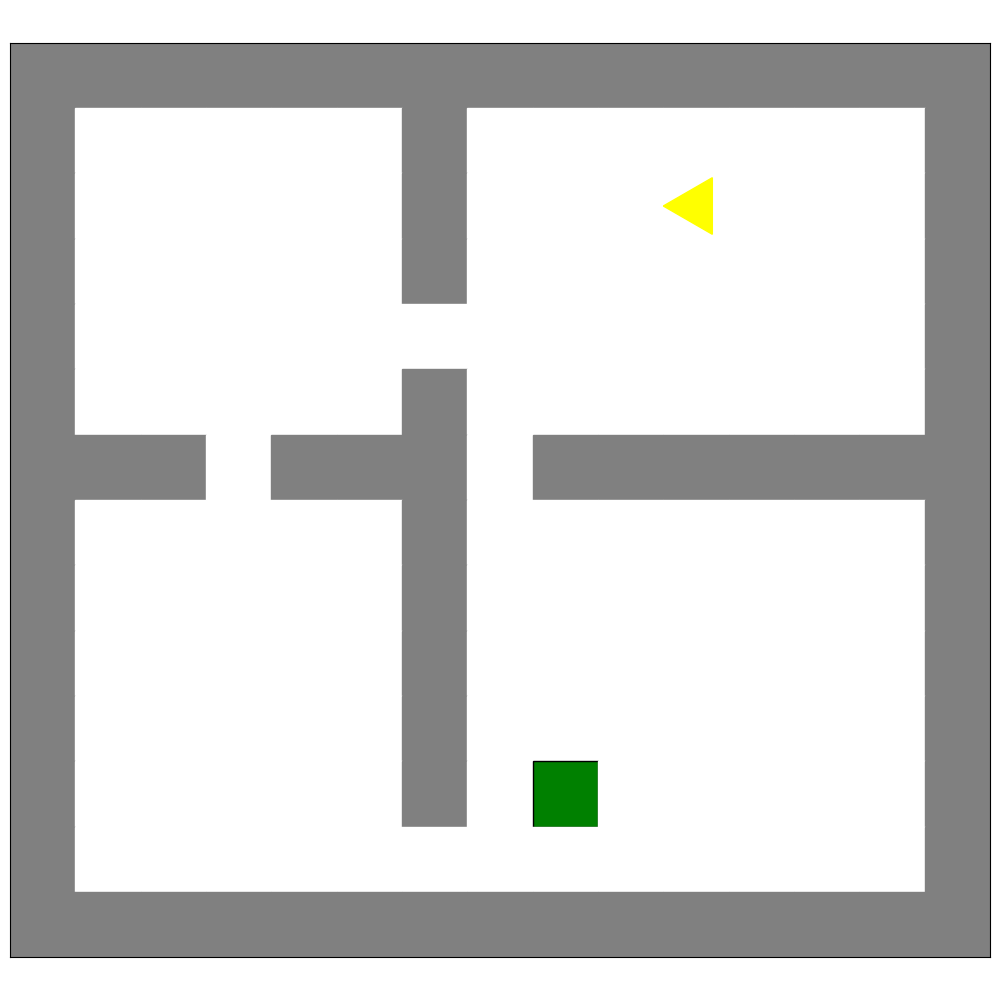}\\
   \includegraphics[width=\linewidth]{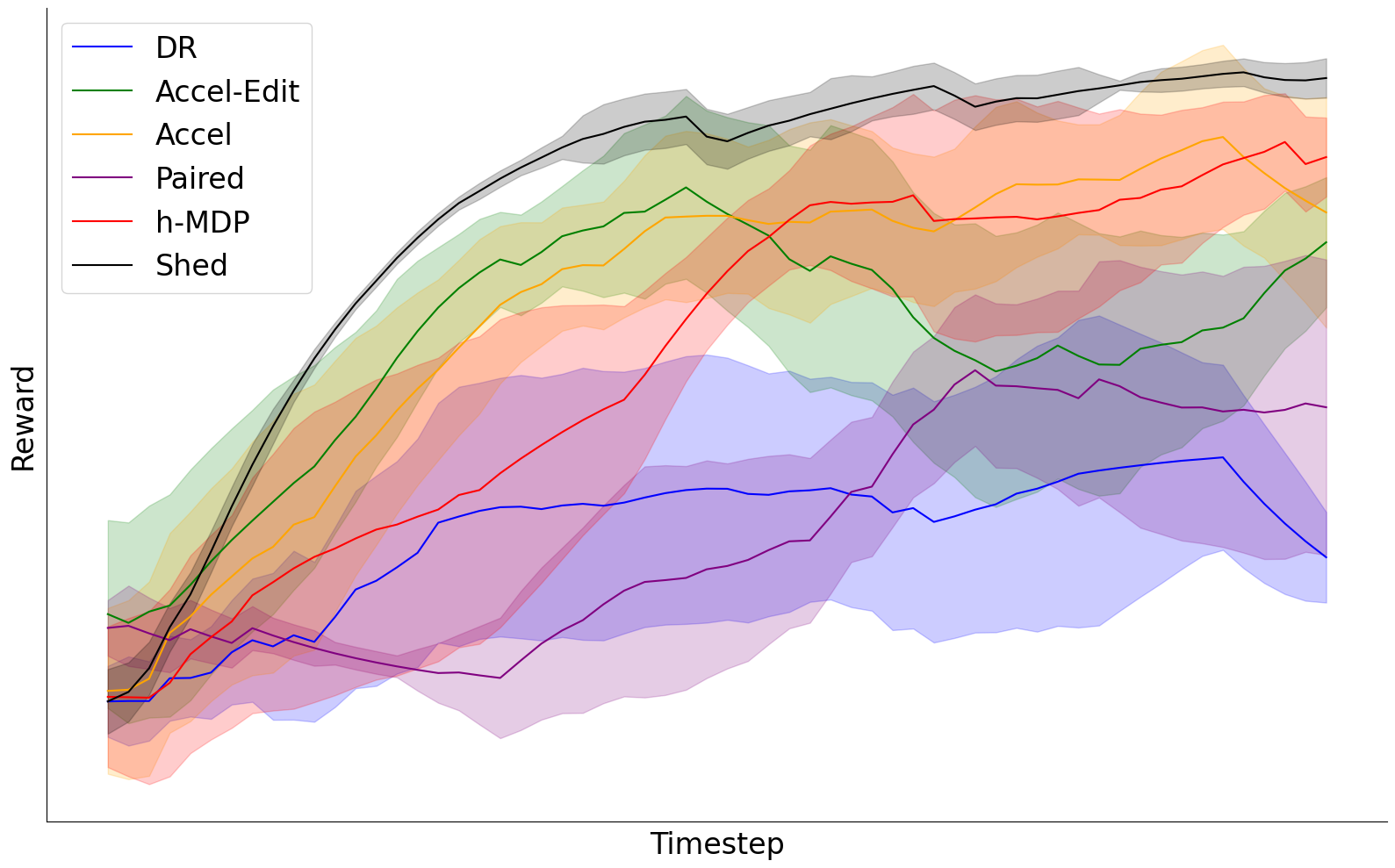}
   \caption{Four rooms maze environment }
   \label{fig:four_rooms_maze}
 \end{minipage}
 \hfill
 \begin{minipage}[b]{0.48\textwidth}
   \centering
   \includegraphics[width=\linewidth]{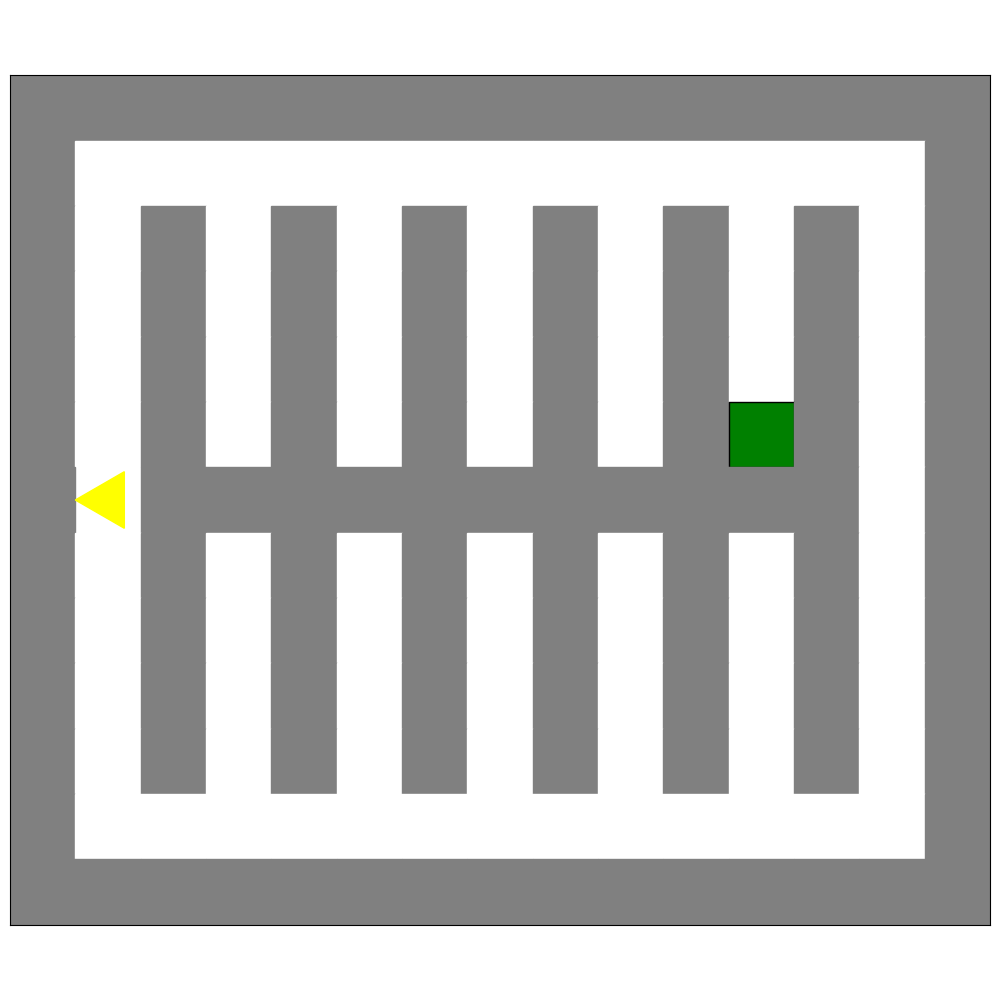}\\
   \includegraphics[width=\linewidth]{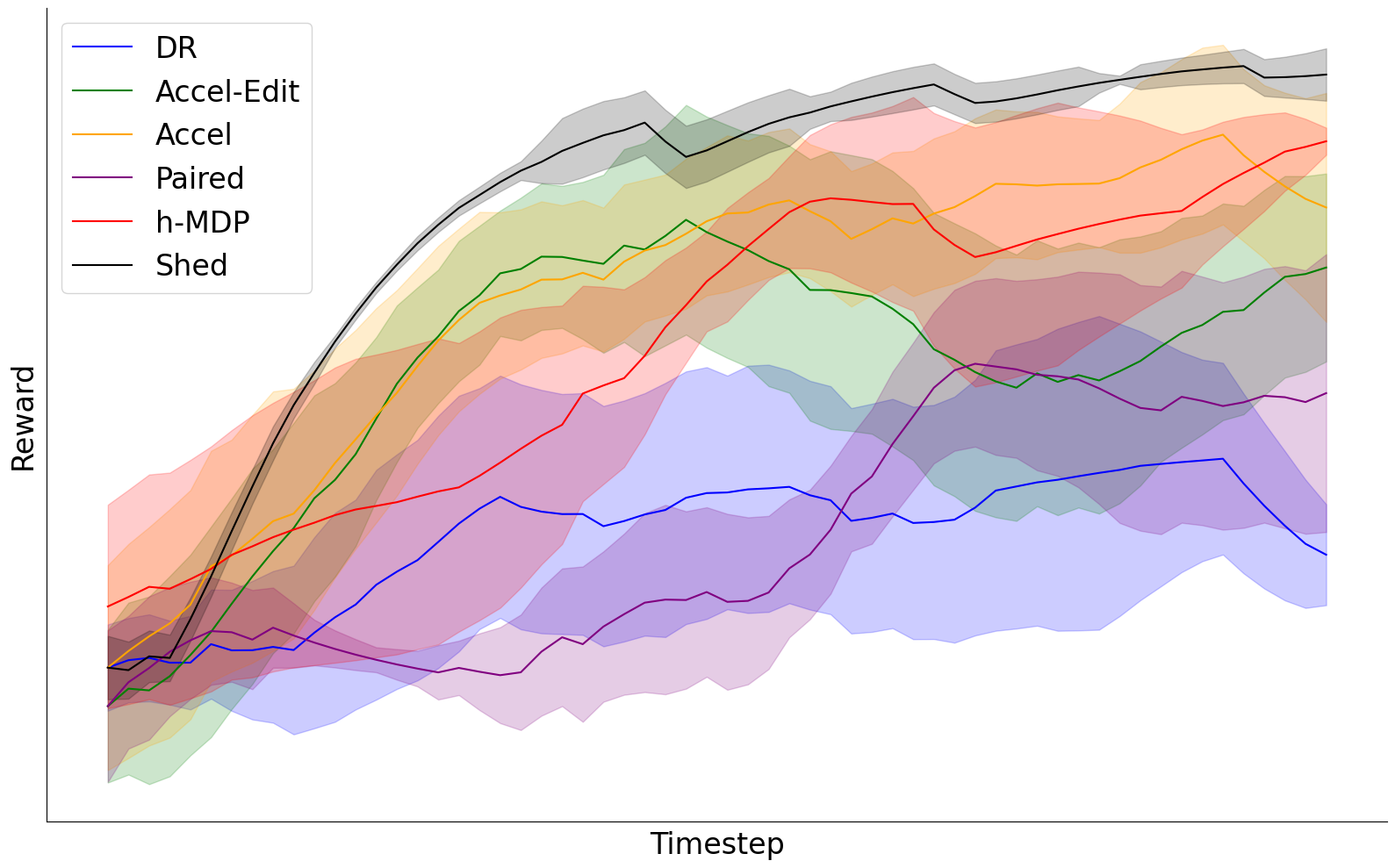}
   \caption{Corridor maze environment }
   \label{fig:corridor_maze}
 \end{minipage}
\end{figure}
\clearpage
\newpage
\section*{NeurIPS Paper Checklist}
\begin{enumerate}

\item {\bf Claims}
    \item[] Question: Do the main claims made in the abstract and introduction accurately reflect the paper's contributions and scope?
    \item[] Answer: \answerYes{} 
    \item[] Justification: We have, to the best of our abilities, presented our domain and motivations clearly in our abstract and throughout our work.
    \item[] Guidelines:
    \begin{itemize}
        \item The answer NA means that the abstract and introduction do not include the claims made in the paper.
        \item The abstract and/or introduction should clearly state the claims made, including the contributions made in the paper and important assumptions and limitations. A No or NA answer to this question will not be perceived well by the reviewers. 
        \item The claims made should match theoretical and experimental results, and reflect how much the results can be expected to generalize to other settings. 
        \item It is fine to include aspirational goals as motivation as long as it is clear that these goals are not attained by the paper. 
    \end{itemize}

\item {\bf Limitations}
    \item[] Question: Does the paper discuss the limitations of the work performed by the authors?
    \item[] Answer: \answerYes{} 
    \item[] Justification: This is done in the final section.
    \item[] Guidelines:
    \begin{itemize}
        \item The answer NA means that the paper has no limitation while the answer No means that the paper has limitations, but those are not discussed in the paper. 
        \item The authors are encouraged to create a separate "Limitations" section in their paper.
        \item The paper should point out any strong assumptions and how robust the results are to violations of these assumptions (e.g., independence assumptions, noiseless settings, model well-specification, asymptotic approximations only holding locally). The authors should reflect on how these assumptions might be violated in practice and what the implications would be.
        \item The authors should reflect on the scope of the claims made, e.g., if the approach was only tested on a few datasets or with a few runs. In general, empirical results often depend on implicit assumptions, which should be articulated.
        \item The authors should reflect on the factors that influence the performance of the approach. For example, a facial recognition algorithm may perform poorly when image resolution is low or images are taken in low lighting. Or a speech-to-text system might not be used reliably to provide closed captions for online lectures because it fails to handle technical jargon.
        \item The authors should discuss the computational efficiency of the proposed algorithms and how they scale with dataset size.
        \item If applicable, the authors should discuss possible limitations of their approach to address problems of privacy and fairness.
        \item While the authors might fear that complete honesty about limitations might be used by reviewers as grounds for rejection, a worse outcome might be that reviewers discover limitations that aren't acknowledged in the paper. The authors should use their best judgment and recognize that individual actions in favor of transparency play an important role in developing norms that preserve the integrity of the community. Reviewers will be specifically instructed to not penalize honesty concerning limitations.
    \end{itemize}

\item {\bf Theory assumptions and proofs}
    \item[] Question: For each theoretical result, does the paper provide the full set of assumptions and a complete (and correct) proof?
    \item[] Answer: \answerYes 
    \item[] Justification: For our discussion on evaluation environments, we present an overview in the main paper, and a full theory discussion in the appendix.
    \item[] Guidelines:
    \begin{itemize}
        \item The answer NA means that the paper does not include theoretical results. 
        \item All the theorems, formulas, and proofs in the paper should be numbered and cross-referenced.
        \item All assumptions should be clearly stated or referenced in the statement of any theorems.
        \item The proofs can either appear in the main paper or the supplemental material, but if they appear in the supplemental material, the authors are encouraged to provide a short proof sketch to provide intuition. 
        \item Inversely, any informal proof provided in the core of the paper should be complemented by formal proofs provided in appendix or supplemental material.
        \item Theorems and Lemmas that the proof relies upon should be properly referenced. 
    \end{itemize}

    \item {\bf Experimental result reproducibility}
    \item[] Question: Does the paper fully disclose all the information needed to reproduce the main experimental results of the paper to the extent that it affects the main claims and/or conclusions of the paper (regardless of whether the code and data are provided or not)?
    \item[] Answer: \answerYes 
    \item[] Justification: We have highlighted all our methods and hyperparameters.
    \item[] Guidelines:
    \begin{itemize}
        \item The answer NA means that the paper does not include experiments.
        \item If the paper includes experiments, a No answer to this question will not be perceived well by the reviewers: Making the paper reproducible is important, regardless of whether the code and data are provided or not.
        \item If the contribution is a dataset and/or model, the authors should describe the steps taken to make their results reproducible or verifiable. 
        \item Depending on the contribution, reproducibility can be accomplished in various ways. For example, if the contribution is a novel architecture, describing the architecture fully might suffice, or if the contribution is a specific model and empirical evaluation, it may be necessary to either make it possible for others to replicate the model with the same dataset, or provide access to the model. In general. releasing code and data is often one good way to accomplish this, but reproducibility can also be provided via detailed instructions for how to replicate the results, access to a hosted model (e.g., in the case of a large language model), releasing of a model checkpoint, or other means that are appropriate to the research performed.
        \item While NeurIPS does not require releasing code, the conference does require all submissions to provide some reasonable avenue for reproducibility, which may depend on the nature of the contribution. For example
        \begin{enumerate}
            \item If the contribution is primarily a new algorithm, the paper should make it clear how to reproduce that algorithm.
            \item If the contribution is primarily a new model architecture, the paper should describe the architecture clearly and fully.
            \item If the contribution is a new model (e.g., a large language model), then there should either be a way to access this model for reproducing the results or a way to reproduce the model (e.g., with an open-source dataset or instructions for how to construct the dataset).
            \item We recognize that reproducibility may be tricky in some cases, in which case authors are welcome to describe the particular way they provide for reproducibility. In the case of closed-source models, it may be that access to the model is limited in some way (e.g., to registered users), but it should be possible for other researchers to have some path to reproducing or verifying the results.
        \end{enumerate}
    \end{itemize}

\item {\bf Open access to data and code}
    \item[] Question: Does the paper provide open access to the data and code, with sufficient instructions to faithfully reproduce the main experimental results, as described in supplemental material?
    \item[] Answer: \answerYes 
    \item[] Justification: The code is provided in supplementary folder.
    \item[] Guidelines:
    \begin{itemize}
        \item The answer NA means that paper does not include experiments requiring code.
        \item Please see the NeurIPS code and data submission guidelines (\url{https://nips.cc/public/guides/CodeSubmissionPolicy}) for more details.
        \item While we encourage the release of code and data, we understand that this might not be possible, so “No” is an acceptable answer. Papers cannot be rejected simply for not including code, unless this is central to the contribution (e.g., for a new open-source benchmark).
        \item The instructions should contain the exact command and environment needed to run to reproduce the results. See the NeurIPS code and data submission guidelines (\url{https://nips.cc/public/guides/CodeSubmissionPolicy}) for more details.
        \item The authors should provide instructions on data access and preparation, including how to access the raw data, preprocessed data, intermediate data, and generated data, etc.
        \item The authors should provide scripts to reproduce all experimental results for the new proposed method and baselines. If only a subset of experiments are reproducible, they should state which ones are omitted from the script and why.
        \item At submission time, to preserve anonymity, the authors should release anonymized versions (if applicable).
        \item Providing as much information as possible in supplemental material (appended to the paper) is recommended, but including URLs to data and code is permitted.
    \end{itemize}

\item {\bf Experimental setting/details}
    \item[] Question: Does the paper specify all the training and test details (e.g., data splits, hyperparameters, how they were chosen, type of optimizer, etc.) necessary to understand the results?
    \item[] Answer: \answerYes 
    \item[] Justification: All experiments are described faithfully.
    \item[] Guidelines:
    \begin{itemize}
        \item The answer NA means that the paper does not include experiments.
        \item The experimental setting should be presented in the core of the paper to a level of detail that is necessary to appreciate the results and make sense of them.
        \item The full details can be provided either with the code, in appendix, or as supplemental material.
    \end{itemize}

\item {\bf Experiment statistical significance}
    \item[] Question: Does the paper report error bars suitably and correctly defined or other appropriate information about the statistical significance of the experiments?
    \item[] Answer: \answerYes 
    \item[] Justification: We provide standard error for all the results reported.
    \item[] Guidelines:
    \begin{itemize}
        \item The answer NA means that the paper does not include experiments.
        \item The authors should answer "Yes" if the results are accompanied by error bars, confidence intervals, or statistical significance tests, at least for the experiments that support the main claims of the paper.
        \item The factors of variability that the error bars are capturing should be clearly stated (for example, train/test split, initialization, random drawing of some parameter, or overall run with given experimental conditions).
        \item The method for calculating the error bars should be explained (closed form formula, call to a library function, bootstrap, etc.)
        \item The assumptions made should be given (e.g., Normally distributed errors).
        \item It should be clear whether the error bar is the standard deviation or the standard error of the mean.
        \item It is OK to report 1-sigma error bars, but one should state it. The authors should preferably report a 2-sigma error bar than state that they have a 96\% CI, if the hypothesis of Normality of errors is not verified.
        \item For asymmetric distributions, the authors should be careful not to show in tables or figures symmetric error bars that would yield results that are out of range (e.g. negative error rates).
        \item If error bars are reported in tables or plots, The authors should explain in the text how they were calculated and reference the corresponding figures or tables in the text.
    \end{itemize}

\item {\bf Experiments compute resources}
    \item[] Question: For each experiment, does the paper provide sufficient information on the computer resources (type of compute workers, memory, time of execution) needed to reproduce the experiments?
    \item[] Answer: \answerYes{} 
    \item[] Justification: Hardware resources are stated in the appendix
    \item[] Guidelines:
    \begin{itemize}
        \item The answer NA means that the paper does not include experiments.
        \item The paper should indicate the type of compute workers CPU or GPU, internal cluster, or cloud provider, including relevant memory and storage.
        \item The paper should provide the amount of compute required for each of the individual experimental runs as well as estimate the total compute. 
        \item The paper should disclose whether the full research project required more compute than the experiments reported in the paper (e.g., preliminary or failed experiments that didn't make it into the paper). 
    \end{itemize}
    
\item {\bf Code of ethics}
    \item[] Question: Does the research conducted in the paper conform, in every respect, with the NeurIPS Code of Ethics \url{https://neurips.cc/public/EthicsGuidelines}?
    \item[] Answer: \answerYes 
    \item[] Justification: We have adhered to the NeurIPS code of ethics.
    \item[] Guidelines:
    \begin{itemize}
        \item The answer NA means that the authors have not reviewed the NeurIPS Code of Ethics.
        \item If the authors answer No, they should explain the special circumstances that require a deviation from the Code of Ethics.
        \item The authors should make sure to preserve anonymity (e.g., if there is a special consideration due to laws or regulations in their jurisdiction).
    \end{itemize}

\item {\bf Broader impacts}
    \item[] Question: Does the paper discuss both potential positive societal impacts and negative societal impacts of the work performed?
    \item[] Answer: \answerYes{} 
    \item[] Justification: We have discussed our work in contribution to several areas, including UED and sim-to-real, but we do not discuss the negative impacts.
    \item[] Guidelines:
    \begin{itemize}
        \item The answer NA means that there is no societal impact of the work performed.
        \item If the authors answer NA or No, they should explain why their work has no societal impact or why the paper does not address societal impact.
        \item Examples of negative societal impacts include potential malicious or unintended uses (e.g., disinformation, generating fake profiles, surveillance), fairness considerations (e.g., deployment of technologies that could make decisions that unfairly impact specific groups), privacy considerations, and security considerations.
        \item The conference expects that many papers will be foundational research and not tied to particular applications, let alone deployments. However, if there is a direct path to any negative applications, the authors should point it out. For example, it is legitimate to point out that an improvement in the quality of generative models could be used to generate deepfakes for disinformation. On the other hand, it is not needed to point out that a generic algorithm for optimizing neural networks could enable people to train models that generate Deepfakes faster.
        \item The authors should consider possible harms that could arise when the technology is being used as intended and functioning correctly, harms that could arise when the technology is being used as intended but gives incorrect results, and harms following from (intentional or unintentional) misuse of the technology.
        \item If there are negative societal impacts, the authors could also discuss possible mitigation strategies (e.g., gated release of models, providing defenses in addition to attacks, mechanisms for monitoring misuse, mechanisms to monitor how a system learns from feedback over time, improving the efficiency and accessibility of ML).
    \end{itemize}
    
\item {\bf Safeguards}
    \item[] Question: Does the paper describe safeguards that have been put in place for responsible release of data or models that have a high risk for misuse (e.g., pretrained language models, image generators, or scraped datasets)?
    \item[] Answer: \answerNA{} 
    \item[] Justification: Our method is highly specific to the RL and UED context, and needs extra effort for misuse. 
    \item[] Guidelines:
    \begin{itemize}
        \item The answer NA means that the paper poses no such risks.
        \item Released models that have a high risk for misuse or dual-use should be released with necessary safeguards to allow for controlled use of the model, for example by requiring that users adhere to usage guidelines or restrictions to access the model or implementing safety filters. 
        \item Datasets that have been scraped from the Internet could pose safety risks. The authors should describe how they avoided releasing unsafe images.
        \item We recognize that providing effective safeguards is challenging, and many papers do not require this, but we encourage authors to take this into account and make a best faith effort.
    \end{itemize}

\item {\bf Licenses for existing assets}
    \item[] Question: Are the creators or original owners of assets (e.g., code, data, models), used in the paper, properly credited and are the license and terms of use explicitly mentioned and properly respected?
    \item[] Answer: \answerYes{} 
    \item[] Justification: All work is original, and baselines methods are referenced to their original authors.
    \item[] Guidelines:
    \begin{itemize}
        \item The answer NA means that the paper does not use existing assets.
        \item The authors should cite the original paper that produced the code package or dataset.
        \item The authors should state which version of the asset is used and, if possible, include a URL.
        \item The name of the license (e.g., CC-BY 4.0) should be included for each asset.
        \item For scraped data from a particular source (e.g., website), the copyright and terms of service of that source should be provided.
        \item If assets are released, the license, copyright information, and terms of use in the package should be provided. For popular datasets, \url{paperswithcode.com/datasets} has curated licenses for some datasets. Their licensing guide can help determine the license of a dataset.
        \item For existing datasets that are re-packaged, both the original license and the license of the derived asset (if it has changed) should be provided.
        \item If this information is not available online, the authors are encouraged to reach out to the asset's creators.
    \end{itemize}

\item {\bf New assets}
    \item[] Question: Are new assets introduced in the paper well documented and is the documentation provided alongside the assets?
    \item[] Answer: \answerNA{} 
    \item[] Justification: We do not release new assets.
    \item[] Guidelines:
    \begin{itemize}
        \item The answer NA means that the paper does not release new assets.
        \item Researchers should communicate the details of the dataset/code/model as part of their submissions via structured templates. This includes details about training, license, limitations, etc. 
        \item The paper should discuss whether and how consent was obtained from people whose asset is used.
        \item At submission time, remember to anonymize your assets (if applicable). You can either create an anonymized URL or include an anonymized zip file.
    \end{itemize}

\item {\bf Crowdsourcing and research with human subjects}
    \item[] Question: For crowdsourcing experiments and research with human subjects, does the paper include the full text of instructions given to participants and screenshots, if applicable, as well as details about compensation (if any)? 
    \item[] Answer: \answerNA{} 
    \item[] Justification: We do not have human subjects
    \item[] Guidelines:
    \begin{itemize}
        \item The answer NA means that the paper does not involve crowdsourcing nor research with human subjects.
        \item Including this information in the supplemental material is fine, but if the main contribution of the paper involves human subjects, then as much detail as possible should be included in the main paper. 
        \item According to the NeurIPS Code of Ethics, workers involved in data collection, curation, or other labor should be paid at least the minimum wage in the country of the data collector. 
    \end{itemize}

\item {\bf Institutional review board (IRB) approvals or equivalent for research with human subjects}
    \item[] Question: Does the paper describe potential risks incurred by study participants, whether such risks were disclosed to the subjects, and whether Institutional Review Board (IRB) approvals (or an equivalent approval/review based on the requirements of your country or institution) were obtained?
    \item[] Answer: \answerNA{} 
    \item[] Justification: We do not have human subjects
    \item[] Guidelines:
    \begin{itemize}
        \item The answer NA means that the paper does not involve crowdsourcing nor research with human subjects.
        \item Depending on the country in which research is conducted, IRB approval (or equivalent) may be required for any human subjects research. If you obtained IRB approval, you should clearly state this in the paper. 
        \item We recognize that the procedures for this may vary significantly between institutions and locations, and we expect authors to adhere to the NeurIPS Code of Ethics and the guidelines for their institution. 
        \item For initial submissions, do not include any information that would break anonymity (if applicable), such as the institution conducting the review.
    \end{itemize}

\item {\bf Declaration of LLM usage}
    \item[] Question: Does the paper describe the usage of LLMs if it is an important, original, or non-standard component of the core methods in this research? Note that if the LLM is used only for writing, editing, or formatting purposes and does not impact the core methodology, scientific rigorousness, or originality of the research, declaration is not required.
    \item[] Answer: \answerNA{} 
    \item[] Justification: LLMs are not in the core method. We do provide the prompts for ChatGPT in the maze environment, in the Appendix.
    \item[] Guidelines:
    \begin{itemize}
        \item The answer NA means that the core method development in this research does not involve LLMs as any important, original, or non-standard components.
        \item Please refer to our LLM policy (\url{https://neurips.cc/Conferences/2025/LLM}) for what should or should not be described.
    \end{itemize}

\end{enumerate}

\end{document}